\documentclass[10pt,twocolumn,letterpaper]{article}

\usepackage[pagenumbers]{wacv} 

\definecolor{wacvblue}{rgb}{0.21,0.49,0.74}
\usepackage[pagebackref,breaklinks,colorlinks,allcolors=wacvblue]{hyperref}

\usepackage{hyperref}
\usepackage{url}
\usepackage{booktabs}
\usepackage{amsfonts}
\usepackage{bm}
\usepackage{nicefrac}
\usepackage{microtype}
\usepackage[table,xcdraw]{xcolor}
\usepackage{graphicx}
\usepackage{amsmath}
\usepackage{subcaption}
\usepackage{enumitem}
\usepackage{amssymb}
\usepackage{multirow}
\usepackage{tikz}
\usepackage{pgfplots}

\definecolor{crimson}{rgb}{0.86, 0.08, 0.24}

\renewcommand{\cline}[1]{\cmidrule{#1}}
\let\cline\cmidrule

\newcommand{\equalcontrib}{\textsuperscript{*}}

\def\wacvPaperID{455} 
\def\confName{WACV}
\def\confYear{2027}

\title{M-PFT: Angular Margin for Calibrated Prompt-Tuning of Fine-Tuned Vision-Language Models\vspace{-10pt}}

\author{Shihab Aaqil Ahamed\equalcontrib$^{1,2}$ \quad Mohamed Ansaff\equalcontrib$^{3}$ \quad Nang Khin Myat Lay Phyu$^{3}$ \quad
Udaya S.K.P. Miriya \\
Thanthrige$^2$ \quad Chamira U. S. Edussooriya$^2$ \quad
Ranga Rodrigo$^2$ \quad
Muhammad Haris Khan\textbf{$^1$}
\\
$^1$Mohamed bin Zayed University of Artificial Intelligence, Abu Dhabi, UAE \\
$^2$ Dept. of Electronic and Telecommunication Engineering, University of Moratuwa, Sri Lanka \\
$^3$Albukhary International University, Alor Setar, Malaysia
\\
{\tt\small shihabaaqilahamed@gmail.com}
}

\begin{document}
\maketitle

\begin{abstract}
Prompt-tuning has emerged as a parameter-efficient technique for adapting large-scale vision-language models (VLMs) to downstream tasks while keeping the pre-trained weights frozen. However, prompt-tuned VLMs often suffer from a base-to-novel dual miscalibration problem, where models typically show underconfidence on base classes and overconfidence on novel classes. Current VLM calibration approaches address this dual miscalibration either separately or jointly, but lack a geometric understanding of the hyperspherical embedding space and thus fail to exploit angular discriminative properties to adjust distorted decision boundaries. To address this, we propose \textbf{M-PFT}, a novel angular-margin approach for prompt tuning that introduces two complementary objectives to promote class discriminability and semantic anchoring. This enlarges the angular margin by 1) minimizing the maximum cosine similarity between tunable class embeddings to reduce embedding collapse and 2) minimizing the angular distance between each tunable embedding and its corresponding zero-shot embedding to preserve semantic geometry. Extensive experiments across 9 prompt-tuning methods with various backbones on 15 diverse datasets show that our approach reduces the average calibration error on novel classes, while remaining competitive on base classes and maintaining comparable accuracy. Further analyses reveal that M-PFT effectively adjusts distorted decision boundaries induced by prompt tuning while preserving the pre-trained embedding space, improving calibration with negligible computational overhead. Our code is available at \href{https://github.com/MB-Shihab-Aaqil-Ahamed/M-PFT/}{\tt M-PFT}.

\end{abstract}
    
\section{Introduction}
\label{sec:intro}

\begin{figure}[t]
\vspace{-10pt}
\centering
\begin{subfigure}{0.49\linewidth}
\centering \includegraphics[width=\linewidth]{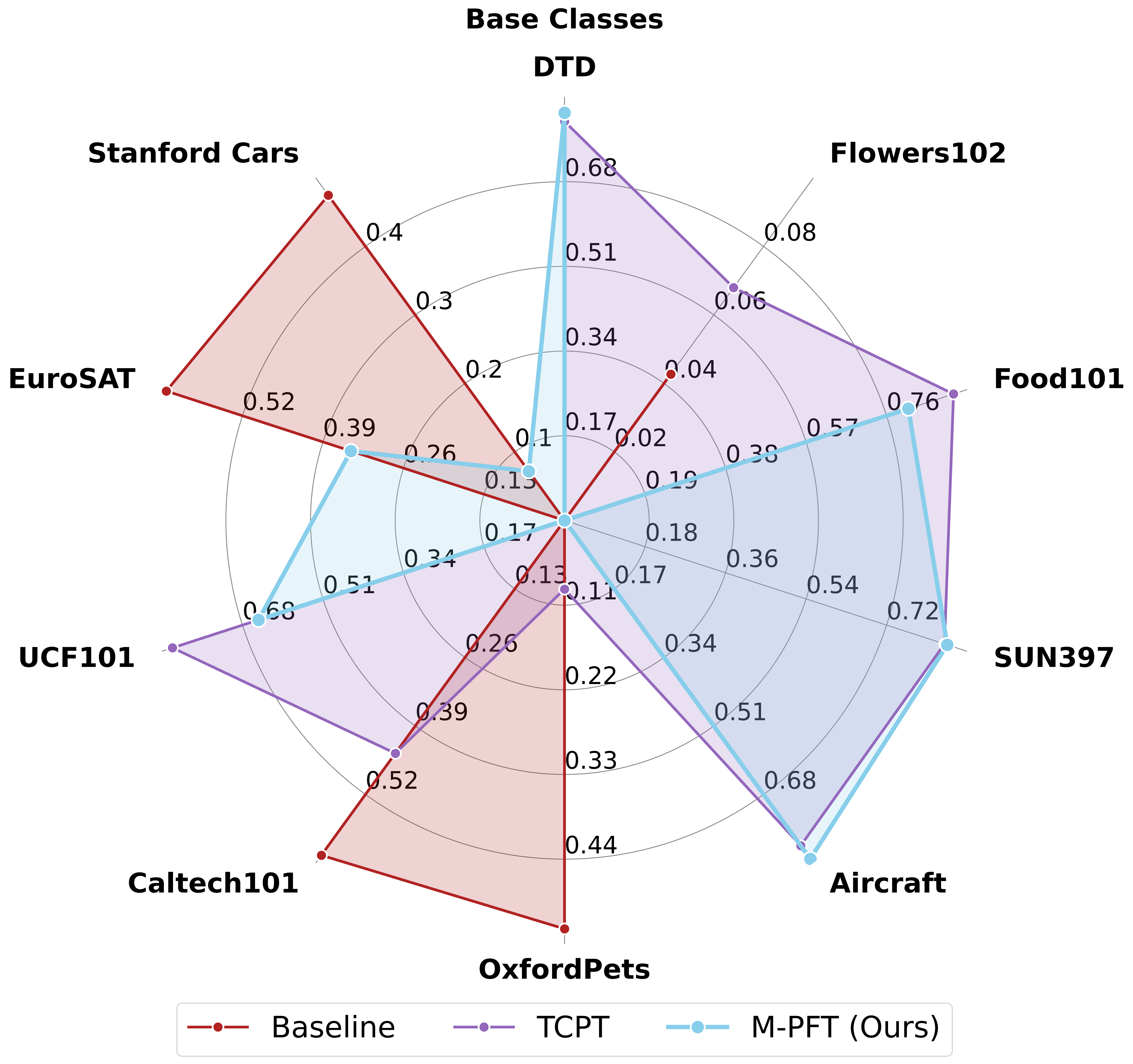}
\end{subfigure}
\hfill
\begin{subfigure}{0.49\linewidth}
\centering
\includegraphics[width=\linewidth]{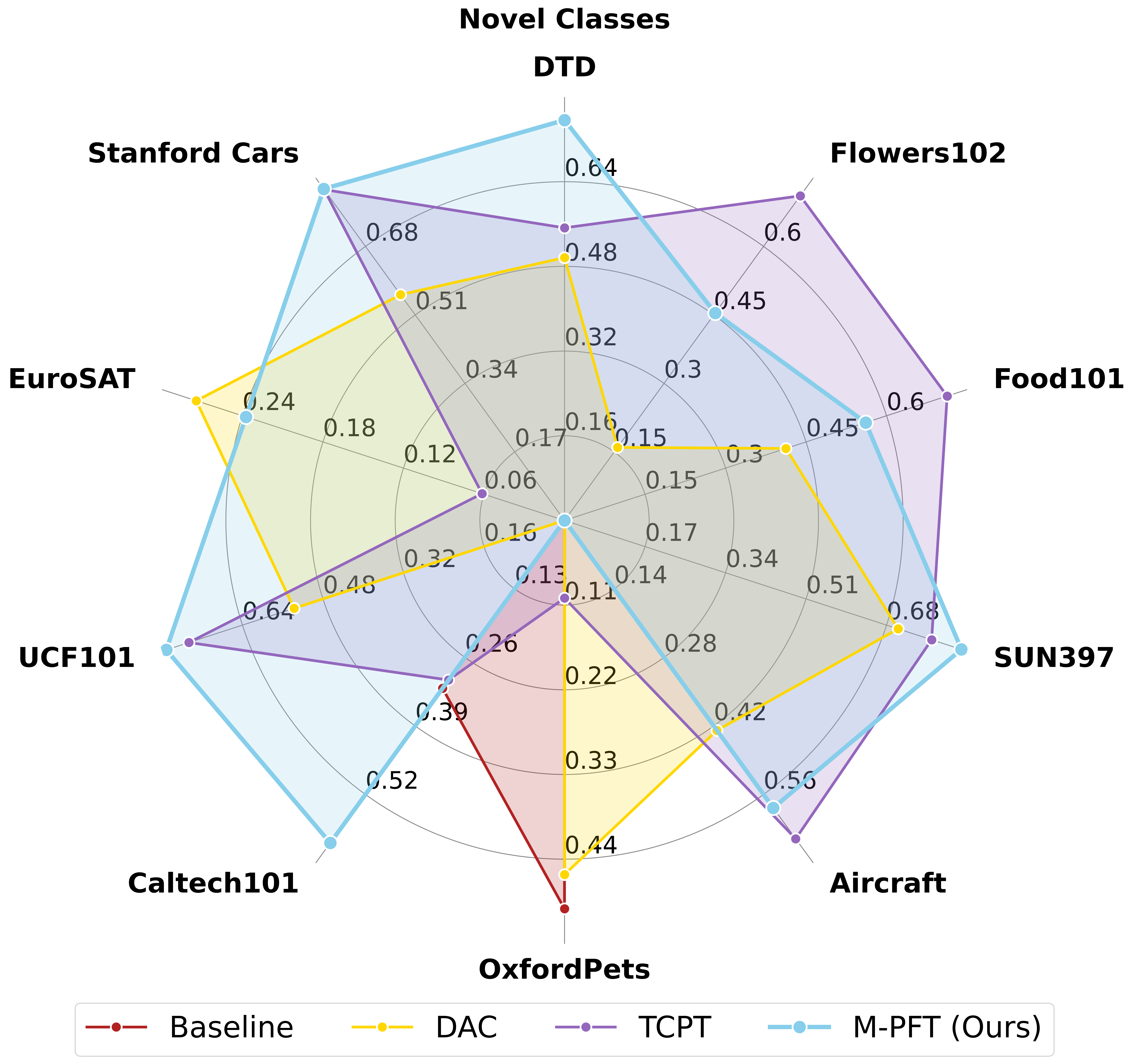}
\end{subfigure}
\vspace{-7pt}
\caption{Comparison of calibration scores with TCPT~\cite{sharifdeen2026towards} and DAC~\cite{wang2024open} on 10 fine-grained classification datasets with CoOp~\cite{zhou2022learning} for base (left) and novel (right) classes. Each axis reports $1 - \frac{\text{ECE}}{\text{ECE}_\text{worst}}$ (percentage of the worst method's ECE removed; larger is better). Ours maximizes the covered area (a higher score indicates a larger footprint), indicating better calibration and effective addressing of the dual miscalibration problem.}
\label{fig:radar}
\vspace{-20pt}
\end{figure}

Foundational vision-language models (VLMs), such as CLIP~\cite{radford2021learning} and ALIGN~\cite{jia2021scaling}, have achieved remarkable progress in open-vocabulary image recognition by leveraging large-scale natural-language supervision. To efficiently adapt such pre-trained large VLMs to domain-specific tasks, where full fine-tuning is often computationally prohibitive, Parameter-Efficient Fine-Tuning (PEFT) techniques have gained considerable attention~\cite{liu2023pre}. Representative PEFT approaches, such as Low-Rank Adaptation~\cite{hu2022lora}, inject trainable low-rank decomposition matrices into transformer layers based on the hypothesis that weight updates have a low intrinsic rank, while Prompt-Tuning~\cite{liu2023pre} optimizes a small subset of learnable prompt tokens while keeping the backbone frozen. Recently, attention has shifted toward Test-Time Prompt-Tuning~\cite{shu2022test}, which, unlike standard prompt-tuning, requires no labeled training data and instead optimizes learnable prompt vectors via gradient descent, adapting them during inference on unlabeled test samples. In open-vocabulary settings, prompt-tuned VLMs can substantially improve performance on base classes while preserving the model's inherent capability to generalize to novel classes~\cite{khattak2023maple,zhou2022conditional}. Such low-cost adaptation is particularly valuable in safety-critical applications such as healthcare and autonomous systems, where full fine-tuning is often infeasible~\cite{zhao2025clip,elhenawy2025vision}.

\begin{figure}[t!]
\vspace{-10pt}
\centering
\includegraphics[width=\linewidth]{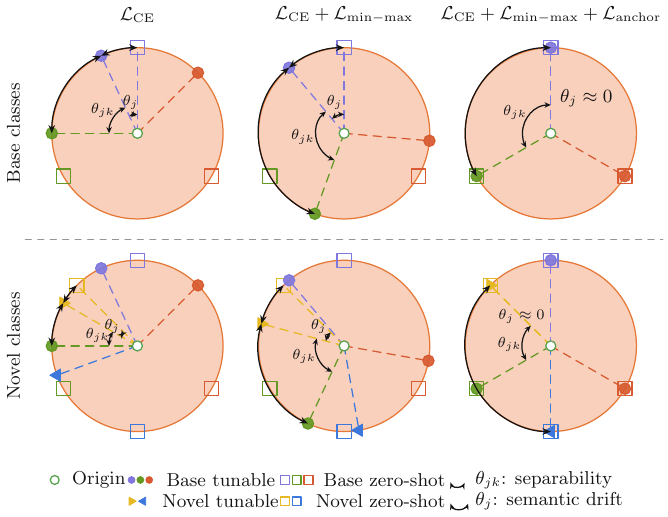}
\vspace{-20pt}
\caption{\textbf{Hyperspherical visualization of the dual miscalibration problem.} $\mathcal{L}_{\text{CE}}$ distorts decision boundaries by inducing embedding collapse $\theta_{jk}\downarrow$ and semantic drift $\theta_{j}\uparrow$. $\mathcal{L}_{\text{min-max}}$ maximizes $\theta_{jk}$ to mitigate collapse and improve class discriminability, while $\mathcal{L}_{\text{anchor}}$ minimizes the drift $\theta_j$ by anchoring tunable embeddings to zero-shot embeddings to preserve semantic geometry. This dual design creates complementary objectives that act synergistically to enlarge the angular margin $\theta_{jk} - \theta_{j}$, yielding well-separated, discriminative decision boundaries while preserving pre-trained semantics and addressing the dual miscalibration.}
\label{fig:sketch}
\vspace{-15pt}
\end{figure}

However, existing few-shot prompt-tuning techniques~\cite{zhou2022learning,zhou2022conditional,yao2023visual,khattak2023maple,lu2022prompt,zhu2023prompt,khattak2023self,zheng2025hierarchical,guo2025mmrl} for VLMs primarily focus on improving classification accuracy while often overlooking reliable confidence calibration, i.e., the degree to which a model's predicted confidence reflects the true likelihood of correctness~\cite{guo2017calibration,wang2023calibration}, which is evaluated with Expected Calibration Error (ECE). For example, prompt-tuned VLMs exhibit base-to-novel dual miscalibration, which typically shows underconfidence on base classes and overconfidence on novel classes~\cite{wang2024understanding,sharifdeen2026towards}. When deployed in safety-critical applications, this dual miscalibration is harmful in two ways: overconfidence often results in models assigning high confidence to incorrect predictions, leading to inappropriate automation or missed human review, while underconfidence assigns low confidence to correct predictions, which is associated with uncertain or unreliable confidence estimates undermining triage, wasting expert time, and eroding trust~\cite{minderer2021revisiting,nixon2019measuring,cao2024deep,zhu2022rethinking,ao2023two}.

\let\thefootnote\relax\footnotetext{\equalcontrib These authors contributed equally to this work.}

To maintain well-calibrated confidence estimates across both base and novel classes, some progress has been made toward addressing dual miscalibration. In light of this, prior works such as DAC~\cite{wang2024open} scale the temperature based on the textual deviation (TD) score between embeddings of each novel class. CAC~\cite{lv2025contrast} adjusts logits based on image-text alignment, while DOR~\cite{wang2024understanding} samples textual outliers that do not overlap with base classes. Another approach~\cite{murugesan2024robust} normalizes logits and constrains them to the original zero-shot VLM distribution by modifying the softmax expression with margin penalization. TCPT~\cite{sharifdeen2026towards} introduces an auxiliary regularization loss to stabilize decision boundaries with a mean-variance margin while preserving semantic dispersion by aligning the first and second moments of text embeddings. Despite these advances, we observe that these statistical approaches fail to adequately exploit the angular discriminative properties of the hyperspherical embedding space when adjusting distorted decision boundaries, which is crucial for resolving dual miscalibration. Specifically, most methods do not impose explicit geometric constraints during prompt tuning, allowing learned embeddings to deviate from the pre-trained embedding space. We define \(\theta_{jk}\) as the angle between tunable class embeddings and \(\theta_j\) as the angle between a tunable embedding and its corresponding zero-shot embedding. This can result in issues like (1) \emph{embedding collapse}, where tunable class embeddings cluster together \(\left(\theta_{jk}\downarrow\right)\), reducing inter-class discriminability~\cite{liu2021learning}, and (2) \emph{semantic drift}, where tunable embeddings move away from their corresponding zero-shot embeddings \(\left(\theta_j\uparrow\right)\), distorting decision boundaries, degrading model's ability to learn semantic relationships between class embeddings, and inducing spurious overconfident predictions on novel inputs~\cite{wang2024understanding,lv2025contrast}. Furthermore, auxiliary losses employed by existing methods target statistical properties rather than angular discriminative properties of class embeddings. This can introduce gradients in opposing directions that partially offset each other, limiting their ability to adjust distorted decision boundaries~\cite{murugesan2024robust,sharifdeen2026towards}. Consequently, these methods cannot effectively enlarge the angular margin and remain suboptimal for calibration.

Inspired by these insights, we propose \textbf{M-PFT}, a novel angular optimization approach that introduces a simple yet effective angular-margin approach into prompt tuning. This exploits the angular discriminative properties of the hyperspherical embedding space~\cite{liu2018learning,liu2021learning} with negligible computational overhead. To address dual miscalibration, our proposed method 1) promotes class discriminability by minimizing the maximum cosine similarity between tunable class embeddings, encouraging discriminative decision boundaries, and 2) enforces semantic anchoring by minimizing the angular distance between each tunable embedding and its corresponding zero-shot embedding, thereby preserving semantic geometry. As shown in Fig.~\ref{fig:sketch}, these complementary objectives act synergistically to enlarge the angular margin \(\left(\theta_{jk} - \theta_j\right)\): \(\mathcal{L}_\text{min-max}\) maximizes \(\theta_{jk}\), while \(\mathcal{L}_{\text{anchor}}\) minimizes \(\theta_j\). This absence of conflicting gradients (see Tab.~\ref{tab:gradient_avg} and Fig.~\ref{fig:gradient}) and yields well-separated, discriminative decision boundaries while preserving the pre-trained embedding space, thereby reducing base-class underconfidence and novel-class overconfidence, as shown in Fig.~\ref{fig:radar}. Our major contributions are summarized as follows:

\begin{itemize}

\item We propose a novel angular-margin approach that exploits the angular discriminative properties of the hyperspherical embedding space to address dual miscalibration of prompt-tuned VLMs, improving upon leading calibration techniques for prompt tuning.

\item We introduce two complementary objectives that adjust distorted decision boundaries while preserving the semantic geometry of the pre-trained embedding space. The angular margin is enlarged by 1) minimizing the maximum cosine similarity between tunable class embeddings and 2) minimizing the angle between each tunable embedding and its corresponding zero-shot embedding.

\item We conduct extensive experiments across \(9\) prompt-tuning methods and \(15\) diverse datasets, showing that M-PFT achieves better calibration without compromising accuracy and outperforms post-hoc and training-time baselines.

\end{itemize}

\section{Related Work}
\label{sec:work}

\textbf{Prompt tuning of large VLMs.}
Early research focused on static prompts shared across all inputs (CoOp~\cite{zhou2022learning}), while subsequent work introduced instance-conditioned prompts that adapt dynamically to each image input (CoCoOp~\cite{zhou2022conditional}). Knowledge-guided methods, such as KgCoOp~\cite{yao2023visual}, regularize the learned prompt space by minimizing the distance between learned and hand-crafted textual embeddings, while gradient-aligned methods, such as ProDA~\cite{lu2022prompt} and ProGrad~\cite{zhu2023prompt}, selectively update prompts whose gradients align with the zero-shot CLIP objective. Recent advances incorporated multimodal prompt learning (MaPLe~\cite{khattak2023maple}), visual context modulation (PromptSRC~\cite{khattak2023self}), bidirectional knowledge flow (HiCroPL~\cite{zheng2025hierarchical}), and modality-agnostic representation spaces (MMRL~\cite{guo2025mmrl}).
These techniques are typically trained with few-shot supervision on a small set of base-class samples that require labeled data, and evaluated for zero-shot generalization to novel classes. This paradigm improves open-vocabulary recognition without fully fine-tuning VLMs. However, while existing methods improve classification accuracy, they often overlook calibrating predicted probabilities, leading to base-to-novel dual miscalibration and failure for reliable deployment.

\textbf{Calibration of deep neural networks.}
Traditional post-hoc techniques, such as Temperature Scaling~\cite{guo2017calibration}, Platt Scaling~\cite{platt1999probabilistic}, and Histogram Binning~\cite{naeini2015obtaining}, adjust confidence estimates after training by modifying logits on a held-out validation set. While effective for in-distribution data, these methods often struggle to generalize to out-of-distribution scenarios because they rely on labeled target-domain data. In contrast, train-time methods incorporate auxiliary losses during training to penalize miscalibrated predictions, including label smoothing~\cite{muller2019does}, focal loss~\cite{mukhoti2020calibrating}, and margin-based calibration~\cite{murugesan2024robust}. Although these approaches improve reliable uncertainty estimates, they typically require fully labeled datasets and full fine-tuning, making them impractical for PEFT techniques such as prompt tuning in few-shot settings.

\textbf{Calibration for prompt-tuned VLMs.}
DAC~\cite{wang2024open} employs post-hoc temperature scaling guided by semantic distances between class embeddings, but may compromise the sharp decision boundaries of pre-trained models~\cite{sharifdeen2026towards}.~\cite{murugesan2024robust} introduced zero-shot normalization and sample-adaptive logit scaling to realign prompt-tuned models with zero-shot CLIP by adjusting expanded logit distributions. Other efforts focus on VLM calibration for test-time prompt tuning: C-TPT~\cite{yoon2024c} maximizes average textual feature dispersion, A-TPT~\cite{ahamed2026atpt} promotes angular diversity, and AGI-TPT~\cite{anonymous2026agitpt} enforces attainable geometric separation for uniformity. MVC-Bench~\cite{sharifdeen2026mvc} evaluates medical imaging calibration and stabilizes decision boundaries with a mean-variance margin, though these approaches introduce additional inference-time overhead. 
These collective studies highlight the importance of class discriminability for reducing base-class underconfidence, preserving the zero-shot CLIP embedding geometry for better novel-class calibration, and the limitations of post-hoc corrections in restoring pre-trained calibration properties. Our proposed method exploits the angular discriminative properties of the embedding space during training to address the observed base-to-novel dual miscalibration pattern without sacrificing discriminative strength.

\section{Proposed Method}
\label{sec:method}

\begin{figure*}[t]
\vspace{-15pt}
\centering
\resizebox{0.98\textwidth}{!}{%
\begin{minipage}{0.22\textwidth} \centering $\mathcal{L}_{\text{CE}}$ \end{minipage}
\begin{minipage}{0.22\textwidth} \centering $\mathcal{L}_{\text{CE}} + \mathcal{L}_{\text{min-max}}$ \end{minipage}
\begin{minipage}{0.22\textwidth} \centering $\mathcal{L}_{\text{CE}} + \mathcal{L}_{\text{min-max}} + \mathcal{L}_{\text{anchor}}$ \end{minipage}
\begin{minipage}{0.3\textwidth} \centering \small Cosine Similarity vs. ECE \end{minipage}%
}

\vspace{1mm}

\resizebox{\textwidth}{!}{%
\rotatebox[origin=l]{90}{\hspace{3.5em} Base class}
\begin{subfigure}{0.22\textwidth}
\includegraphics[width=\textwidth]{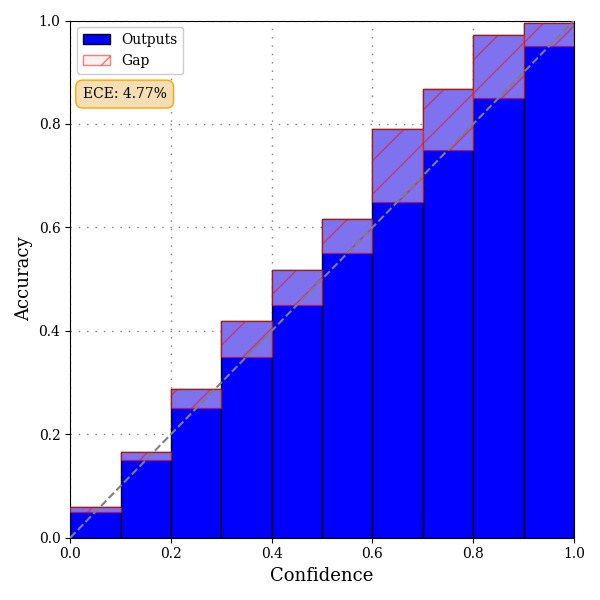}
\end{subfigure}
\begin{subfigure}{0.22\textwidth}
\includegraphics[width=\textwidth]{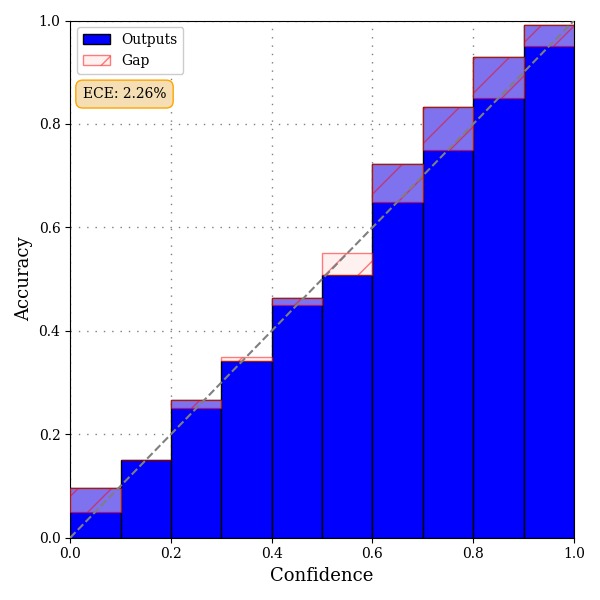}
\end{subfigure}
\begin{subfigure}{0.22\textwidth}
\includegraphics[width=\textwidth]{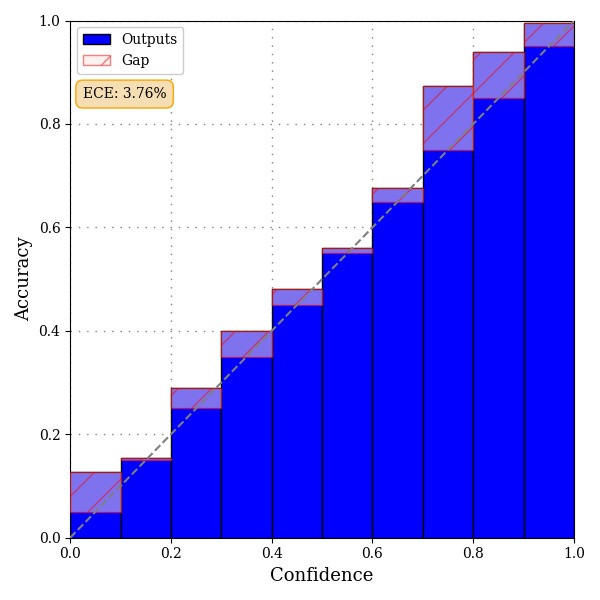}
\end{subfigure}
\begin{subfigure}{0.3\textwidth}
\includegraphics[width=\textwidth]{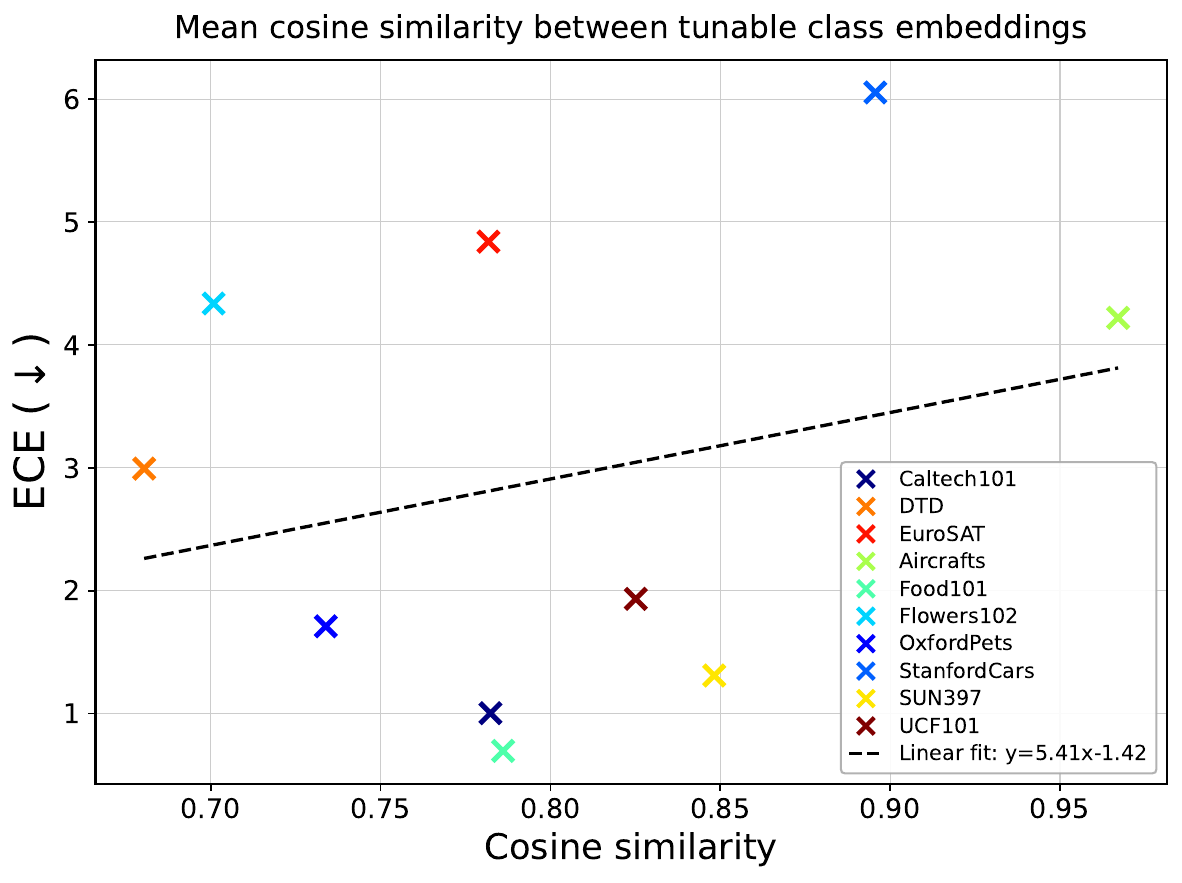}
\end{subfigure}%
}

\vspace{2mm}

\resizebox{0.98\textwidth}{!}{%
\rotatebox[origin=l]{90}{\hspace{3.5em} Novel class}
\begin{subfigure}{0.22\textwidth}
\includegraphics[width=\textwidth]{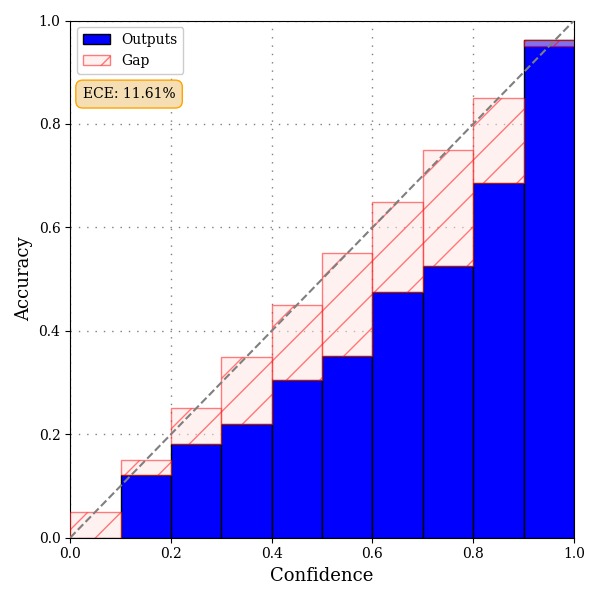}
\end{subfigure}
\begin{subfigure}{0.22\textwidth}
\includegraphics[width=\textwidth]{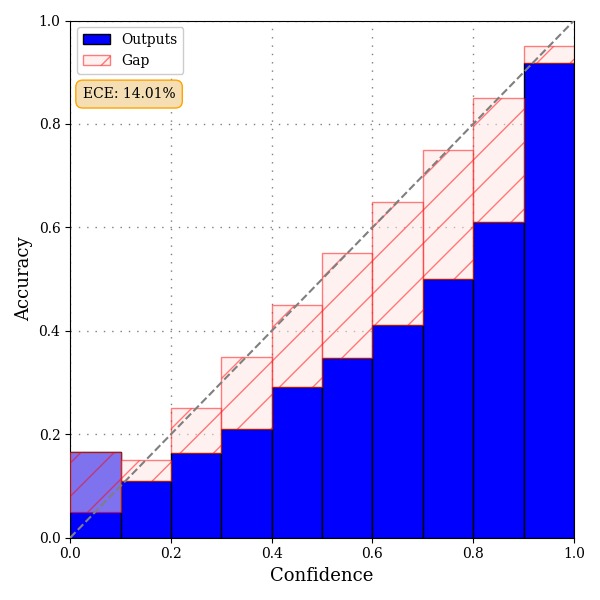}
\end{subfigure}
\begin{subfigure}{0.22\textwidth}
\includegraphics[width=\textwidth]{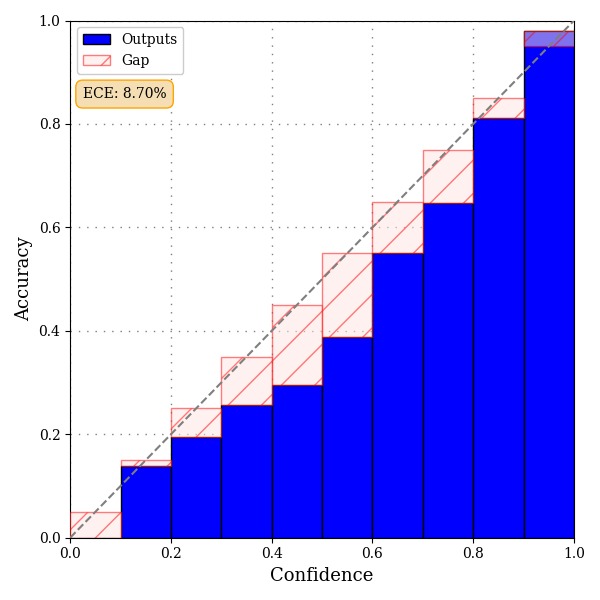}
\end{subfigure}
\begin{subfigure}{0.3\textwidth}
\includegraphics[width=1.01\textwidth]{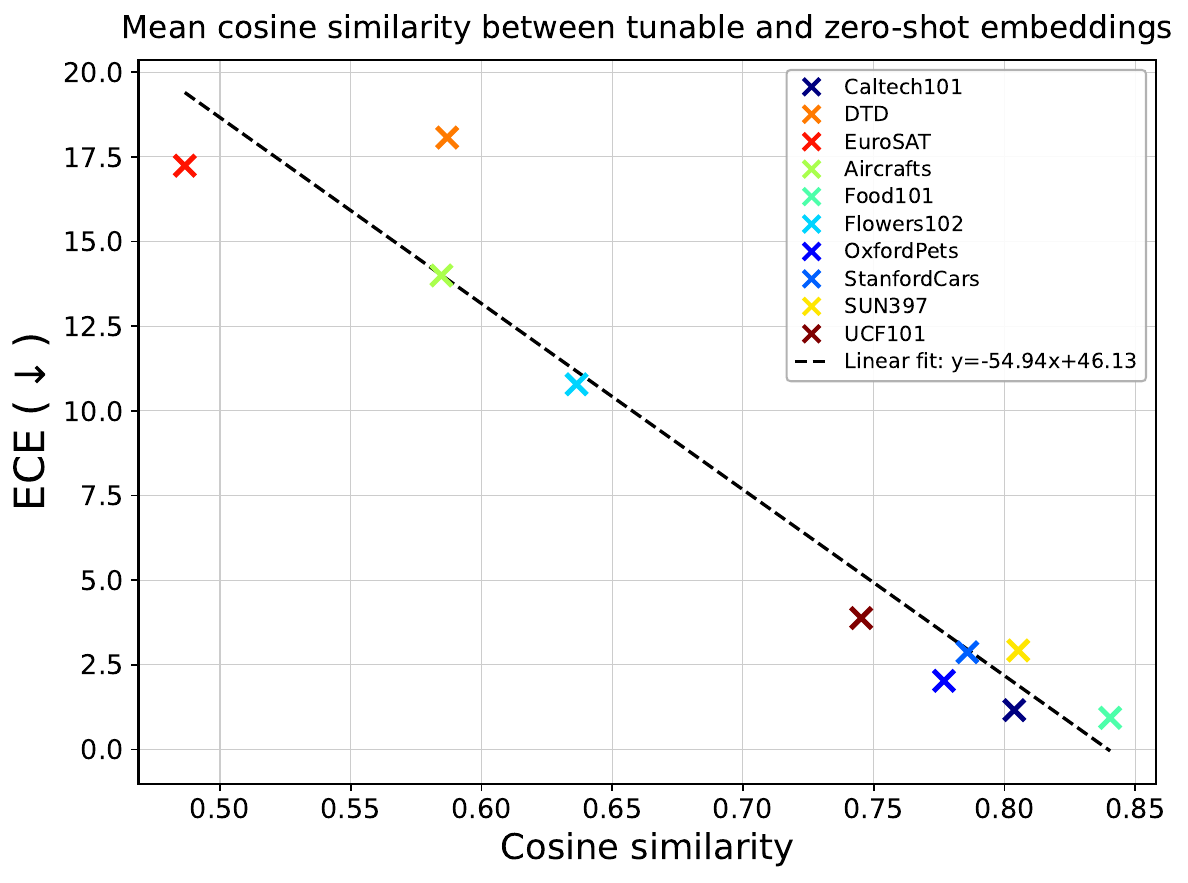}
\end{subfigure}%
}
\vspace{-7pt}
\caption{\textbf{Analyses of dual miscalibration of prompt-tuned VLMs.} Base classes show underconfidence mitigated by our complementary objective, whereas novel classes show overconfidence alleviated by the proposed method. The cross-dataset cosine similarity analyses reveal a negative correlation between angular margin (\(\theta_{jk} - \theta_j\)) and ECE, where enlarging the margin lowers calibration error. This shows prompt tuning distorts decision boundaries with embedding collapse (\(\theta_{jk} \downarrow\)) and semantic drift (\(\theta_j \uparrow\)), resulting in base-to-novel dual miscalibration. These conflicting trends motivate the angular margin framework's class discriminability and semantic anchoring objectives.}
\label{fig:dual_miscalibration}
\vspace{-10pt}
\end{figure*}

\subsection{Preliminaries}

\textbf{Zero-shot inference with prompt-tuned VLMs.} CLIP learns a shared multimodal embedding space for images and natural language descriptions through large-scale contrastive pre-training, enabling zero-shot classification without task-specific supervision. CLIP consists of two encoders: an image encoder \(f_i: \mathcal{X} \to \mathbb{R}^{d}\) and a text encoder \(f_t: \mathcal{T} \to \mathbb{R}^{d}\), which map visual and textual inputs into corresponding feature space vectors, respectively. Given an input image \(x \in \mathbb{R}^{H\times W\times C}\), the image encoder extracts feature embedding \(\mathbf{v}=f_i(x)\). Similarly, the text encoder maps a textual prompt \({\tt p}\) associated with a class label \(C\) into a feature vector \(\mathbf{z}=f_t({\tt p})\). During zero-shot inference, each class label \(\mathcal{C} \in C =\{c_k\}_{k=1}^{N}\) is converted into a text prompt using a fixed template, such as \({\tt p}_k = {\tt ``a~photo~of~a~[CLS}_k{\tt ]''}\), and encoded as \(\mathbf{z} = \{\mathbf{z}_k\}_{k=1}^{N}\), where \(\mathbf{z}_k=f_t({\tt p}_k)\). The similarity between the image embedding \(\mathbf{v}\) and each text embedding \(\mathbf{z}_k\) is computed using cosine similarity, \(s_k=\text{sim}(\mathbf{v}, \mathbf{z}_k)\). The predicted class probabilities are then computed with a softmax function as \(\mathbb{P}(c_k|x)=\frac{\exp (\tau s_k)}{\sum_{i=1}^{N}\exp (\tau s_{i})}\), where \(\tau\) is a temperature parameter. The predicted label \(\hat{c{}}\) and confidence are \(\hat{c}=\arg\max_k\mathbb{P}(c_k|x)\) and \(\hat{p}=\max_k\mathbb{P}(c_k|x)\), respectively.

\textbf{Prompt tuning of fine-tuned VLMs.} While zero-shot classification with handcrafted templates is widely adopted, it often requires domain-specific heuristics and may not be optimal across a variety of tasks. To address this, prompt tuning optimizes learnable prompt tokens to adapt VLMs to domain-specific tasks. Instead of relying on hand-crafted prompt templates, a set of learnable context tokens \(\mathcal{T}_p = {{\tt l}_1,\ldots, {\tt l}_M}\), where \(M\) is the context length, is introduced as a prefix to the class name. For class \(c_k \in \mathcal{C}\), the prompt is defined as \({\tt p}_k = [{\tt l}_1,\ldots, {\tt l}_M, {\tt e}_k]\), where \({\tt e}_k\) is the static class-name embedding. The text encoder maps it into a class representation \(\mathbf{\phi}_k = f_t({\tt p}_k)\). Given image $x$, the cosine similarity for class \(c_k\) is computed as \(s_k = \text{sim}(f_i(x), \mathbf{\phi}_k)\), and class probabilities are obtained via softmax as in zero-shot inference. The prompt tokens \(\mathcal{T}_p\) are optimized using a small labeled training set \(\mathcal{D} = \{(x_j, y_j)\}_{j=1}^{\mathcal{B}}\), typically by minimizing cross-entropy loss over predicted class probabilities. During prompt tuning, the model is trained only on the base classes in a few-shot setting, denoted as \(\mathcal{C}_{\text{base}} \subset \mathcal{C}\), and evaluated on remaining base class samples and novel classes encountered at test time as \(\mathcal{C}_{\text{novel}} = \mathcal{C} \setminus \mathcal{C}_{\text{base}}\). In this section, we first analyze the base-to-novel dual miscalibration problem in prompt-tuned VLMs and then present our training-time method to address it.

\textbf{Dual miscalibration analyses.} To examine the calibration behavior of prompt-tuned VLMs, we conduct a systematic analysis across multiple prompt configurations and datasets. As shown in Fig.~\ref{fig:dual_miscalibration}, reliability diagrams and cosine similarity analyses reveal a dual miscalibration pattern indexed by the angular margin \(\theta_{jk} - \theta_j\). For base classes, reliability plots show underconfidence: predicted probabilities lag behind actual accuracy, reflecting reduced \(\theta_{jk}\) between tunable class embeddings. Fig.~\ref{fig:dual_miscalibration} (upper right), lower cosine similarity corresponds to lower calibration error, where maximizing \(\theta_{jk}\) enlarges the angular margin for better calibration, showing a negative correlation between angular margin and underconfidence. For novel classes, the opposite pattern emerges: predictions are overconfident, assigning high predicted probabilities to incorrect classes, which is exacerbated when tunable embeddings drift from their corresponding zero-shot embeddings \(\left(\theta_j\uparrow\right)\), distorting decision boundaries at test-time. Fig.~\ref{fig:dual_miscalibration} (lower right), higher cosine similarity is associated with lower ECE, where minimizing \(\theta_j\) enlarges the angular margin for better calibration, which shows a negative correlation between angular margin and overconfidence. The scatter plot in Fig.~\ref{fig:dual_miscalibration} (right) quantifies these trends and reveals a negative correlation between angular margin and ECE. Although cross-dataset correlation for the base class is weak, further analyses in Tab.~\ref{tab:gradient_avg} show that increased \(\theta_{jk}\) reduces base ECE to 2.49, supporting the correlation.
These findings show that prompt tuning distorts the well-calibrated decision boundaries of zero-shot CLIP. A well-calibrated open-vocabulary VLM should enlarge the angular margin while preserving the pre-trained embedding space, where increasing \(\theta_{jk}\) reduces base-class underconfidence, while decreasing \(\theta_j\) suppresses novel-class overconfidence. This motivates our angular-margin approach to prevent embedding collapse and preserve semantic geometry. 

\subsection{Proposed Approach: Angular Margin}
\label{sec:angularmargin}

The primary objective of M-PFT is to improve confidence calibration for prompt-tuned discriminative VLMs, such as CLIP, with reliable confidence estimates for base-to-novel dual miscalibration. As shown in Fig.~\ref{fig:sketch}, this miscalibration arises because prompt tuning distorts the decision boundaries of the pre-trained embedding space, making them asymmetric and poorly separated. Specifically, base classes suffer from embedding collapse \(\left(\theta_{jk}\downarrow\right)\), leading to underconfidence, while novel classes suffer from semantic drift \(\left(\theta_j\uparrow\right)\), causing overconfidence. Unlike existing calibration approaches~\cite{murugesan2024robust, wang2024open}, which fail to exploit the angular discriminative properties of the hyperspherical embedding space. Our method addresses this dual miscalibration with two complementary objectives that enlarge the angular margin by 1) \emph{class discriminability}, which promotes discriminative decision boundaries, and 2) \emph{semantic anchoring}, which preserves the semantic geometry of the pre-trained embedding space. 

\subsection{Min-Max Regularization.}
\label{sec:min-max}
Given a tunable class text embedding matrix \(\mathbf{\Phi} = [\mathbf{\phi}_1, \mathbf{\phi}_2, \dots, \mathbf{\phi}_N]^T \in \mathbb{R}^{N \times d}\), where \(\mathbf{\phi}_k \in \mathbb{R}^{d}\) is the embedding for class \(c_k\), \(N\) denotes the number of classes, and \(d\) is the embedding dimension, we normalize each embedding as \(\text{s.t.}~{\forall_j}~\hat{\mathbf{\phi}}_j=\frac{\mathbf{\phi}_j}{{\Vert{}\mathbf{\phi}_j\Vert{}}_2}\)
to construct \(\mathbf{\hat{\Phi}}\). For a predicted class \(c_{j}\) and runner-up class \(c_{k}\), the log-ratio of their probabilities is given by \(\log(p_j / p_k) = \tau v^\top (\mathbf{\phi}_j - \mathbf{\phi}_k)\). Since \(\Vert{}\mathbf{\phi}_j - \mathbf{\phi}_k\Vert{} = 2 \sin(\theta_{jk}/2)\), applying the Cauchy–Schwarz inequality yields \(\vert{}\log(p_j / p_k)\vert{} \le 2\tau \sin(\theta_{jk}/2)\). We use this bound to motivate our approach: when inter-class embedding collapse drives \(\theta_{jk} \to 0\), it shrinks the maximum attainable margin between the top-1 and runner-up classes. This caps the top-1 confidence regardless of sample separability, causing base underconfidence. Well-separated class embeddings yield discriminative decision boundaries, reducing base-class underconfidence. Therefore, we propose min-max regularization to promote class discriminability 
by maximizing angular distance between tunable class text embeddings. Since cosine similarity and angular distance are monotonically inversely related on the unit sphere, we implement this 
by minimizing the cosine similarity between class embeddings. We then compute \(
\bm{S} = \mathbf{\hat{\Phi}}\mathbf{\hat{\Phi}}^T \in \mathbb{R}^{N \times N}\),
where \(\bm{S}_{jk} = \mathbf{\hat{\phi}}_j \cdot \mathbf{\hat{\phi}}_k\) is the cosine similarity between the \(j\)-th and \(k\)-th tunable class embeddings. Since computing the maximum cosine similarity \(\max_{k,~k \neq j} \bm{S}_{jk}\) is computationally expensive, we compute the mean of each embedding’s maximum cosine similarity. To well-separate class embeddings on the unit hypersphere, we penalize the minimally separated pairs by minimizing their maximum cosine similarity: \begin{equation}
\mathcal{L}_{\text{min-max}}=\frac{1}{N} \sum_{j=1}^{N} \max_{k,~k \neq j} \bm{S}_{jk}.
\label{eq:min-max}
\end{equation} 

\subsection{Anchoring Regularization.}
\label{sec:anchor}
Let \(\mathbf{Z}~=~\left[\mathbf{z}_1, \ldots, \mathbf{z}_N\right]^T \in \mathbb{R}^{N \times d}\) denote zero-shot embedding matrix, \(\mathbf{z}_k \in \mathbb{R}^{d}\) corresponds to class \(c_k\), \(N\) is the number of classes, and \(d\) is the embedding dimension. We normalize each embedding as \(\text{s.t.}~\forall_j~\hat{\mathbf{z}}_j=\frac{\mathbf{z}_j}{{\Vert{}\mathbf{z}_j\Vert{}}_2}\)
to construct \(\mathbf{\hat{Z}}\). While min-max regularization shapes confidence near decision boundaries, it operates on tunable class embeddings and does not constrain them within the pre-trained embedding space; their logits are bounded by $|\tau v^\top \mathbf{\phi}_j - \tau v^\top \mathbf{z}_j| \le 2\tau \sin(\theta_j/2)$. During training, \(\mathcal{L}_\text{CE}\) tunes learnable context tokens, which increases \(\mathbf{v}^\top \phi_j\) in the base classes. Novel classes then apply these tuned tokens at test time. However, as semantic drift drives \(\theta_j \uparrow\), the margin inflates, driving calibration error upward and yielding spurious confidence spikes in novel classes. In practice, maximizing angular distance can cause tunable embeddings to drift away from their corresponding zero-shot embeddings, disrupting semantic geometry and amplifying overconfidence in novel classes. As shown in Fig.~\ref{fig:dual_miscalibration} and Tab.~\ref{tab:gradient_avg}, \(\mathcal{L}_{\text{min-max}}\) alone can increase ECE on novel classes. To preserve well-calibrated generalization, we anchor tunable embeddings to their corresponding zero-shot embeddings by minimizing the angle between them, preserving rich semantics. We then compute the alignment matrix \(\bm{A} = \mathbf{\hat{\Phi}}\mathbf{\hat{Z}}^T \in \mathbb{R}^{N \times N}\),
whose diagonal entries \(\bm{A}_{j}\) encode the cosine similarity between each tunable embedding and its corresponding zero-shot embedding. We convert these diagonal entries to angular distances \(
\bm{\theta}_j =\arccos (\bm{A}_j)\),
where \(\bm{\theta}_j\) quantifies the semantic drift for class \(c_j\). We minimize this drift with
\begin{equation}
\mathcal{L}_{\text{anchor}}=\frac{1}{N}\sum\limits_{j=1}^{N} \bm{\theta}^2_{j}.
\label{eq:anchor}
\end{equation} 
where input-dependent prompts are regularized on the per-instance prototype matrix and averaged over the batch, with \(\mathcal{L}_\text{anchor}\) targeting the same fixed zero-shot embeddings.

\textbf{Training objective.} To address dual miscalibration, we combine the two regularization terms described in Sec.~\ref{sec:min-max} and~\ref{sec:anchor} with standard cross-entropy. Here, \(\mathcal{L}_{\text{min-max}}\) promotes discriminability, while \(\mathcal{L}_{\text{anchor}}\) preserves the semantic geometry of the pre-trained embedding space. Together, these objectives enable prompt-tuned models to inherit the well-calibrated behavior observed in zero-shot VLMs.
\begin{equation}
\mathcal{L}_{\text{total}}=\mathcal{L}_{\text{CE}}+\lambda_{\text{1}}\mathcal{L}_{\text{min-max}}+\lambda_{\text{2}}\mathcal{L}_{\text{anchor}},
\end{equation}
where \(\lambda_{\text{1}},~\lambda_{\text{2}} \ge 0\) control the relative importance of each term. The class discriminability and semantic anchoring objective addresses dual miscalibration by reducing base-class underconfidence and novel-class overconfidence while maintaining domain-specific task-adaptation capabilities.

\textbf{Complementary objective.} 
\(\mathcal{L}_{\text{min-max}}\) maximizes \(\theta_{jk}\) by pushing tunable class embeddings apart to enforce class discriminability and reduce base-class underconfidence, while \(\mathcal{L}_{\text{anchor}}\) minimizes \(\theta_j\) by anchoring each tunable embedding close to its zero-shot counterpart to preserve semantic geometry and suppress novel-class overconfident predictions. \(\mathcal{L}_{\text{min-max}}\) alone can drift the tunable embeddings from their corresponding zero-shot embeddings, shrinking the angular margin and amplifying novel-class overconfidence, while \(\mathcal{L}_{\text{anchor}}\) alone leaves \(\theta_{jk}\) unconstrained, failing to address base-class underconfidence. This dual design of M-PFT with gradient blocking creates two complementary objectives that act synergistically to enlarge the angular margin and yield well-separated, discriminative boundaries without sacrificing discriminative strength (see Fig.~\ref{fig:gradient} and Tab.~\ref{tab:gradient_avg}). 

\textbf{Gradient blocking analyses.} The gradient of anchoring regularization w.r.t. zero-shot embedding is blocked in Eq.~\ref{eq:anchor}. Without gradient blocking, zero-shot embeddings would also be pulled toward their corresponding tunable embeddings, resulting in tunable embeddings clustering together and reducing class discriminability (\(\theta_{jk}\downarrow\), as shown in Fig.~\ref{fig:gradient}. This collapse in class discriminability can increase ECE, as confirmed by the analyses in Sec.~\ref{sec:ablation}. But gradient blocking is not always necessary to maintain class discriminability, since the variant without it has lower ECE on some datasets.

\textbf{Numerical stability analyses.} \(\arccos(x)\) can suffer from instability and gradient explosion if its input gets too close to the ends or leaks outside the boundaries of \([-1, 1]\), as its gradient is \(\frac{\partial }{\partial x}\arccos ^{2}(x)=2\cdot \arccos (x)\cdot \frac{\partial }{\partial x}(\arccos (x))=\frac{-2\cdot \arccos (x)}{\sqrt{1-x^{2}}}\).
In Eq.~\ref{eq:anchor}, where \(\theta_{j}\) is computed by \(arccos\) of cosine similarities (the dot product of normalized embeddings \(\phi_j\cdot\mathbf{z}_j^T\)). We therefore clamp it within \(\left(-1, 1\right)\) to the interval \(\left(-1 + 1e-7, 1 - 1e-7\right)\) before computing $\arccos$. This prevents \({\tt NaNs}\) in the $\arccos$ calculation and infinite gradients at \(\pm1\) during backpropagation.

\section{Experiments}
\label{sec:experiments}

The suppl. provides the evaluation setup, datasets, compared methods, implementation details, and evaluation metrics. 

\subsection{Results} 

Tab.~\ref{tab:results_avg} reports the average accuracy and ECE across 10 fine-grained classification datasets for 9 prompt-tuning methods on base and novel classes using CLIP and SigLIP backbones. We also evaluate using other calibration metrics, including MCE and ACE (see suppl. for details). 

\textbf{Calibration performance on base classes.} As shown in Tab.~\ref{tab:results_avg}, M-PFT reduces base-class miscalibration across backbones while preserving accuracy. Averaged across \(10\) datasets, M-PFT lower ECE from \(4.54\) to \(3.85\) on CLIP and from \(2.75\) to \(2.54\) on SigLIP, while maintaining or slightly improving accuracy. The gains are larger for the cross-entropy baseline highly miscalibrated: for CoOp on CLIP, ECE reduces from \(6.82\) to \(2.94\), while raising accuracy from \(81.54\) to \(82.97\). For already well-calibrated methods such as MMRL, calibration remains stable (\(1.60\) ECE). Compared to competitive TCPT, M-PFT achieves the lowest average ECE on SigLIP \((2.63\) to \(2.54)\). M-PFT trails TCPT on base-class ECE with CLIP backbones while retaining higher accuracy on CLIP. These results show that enforcing class discriminability mitigates embedding collapse and resulting base-class underconfidence. 


\textbf{Calibration performance on novel classes.} We evaluate M-PFT on novel classes under open-vocabulary generalization using CLIP and SigLIP backbones. While prompt tuning improves zero-shot accuracy (74.30 Acc., 4.43 ECE on CLIP), it degrades calibration for overconfident methods such as CoOp (12.63 ECE) and ProDA (9.03 ECE). M-PFT significantly improves these miscalibrated methods, reducing ECE for CoOp to 4.75, ProDA to 3.99, and MaPLe from 6.18 to 4.08 while boosting accuracy. Well-calibrated methods like MMRL remain stable (3.72 to 3.47 ECE). We attribute this to anchoring tunable embeddings to the pre-trained geometry, which locally corrects semantic drift. In contrast, post-hoc DAC only rescales confidence, partially aiding MaPLe (6.18 to 4.86 ECE) but overcorrecting MMRL (3.72 to 4.95 ECE). Overall, M-PFT achieves the lowest average ECE (3.86) and highest accuracy (75.32) on CLIP, outperforming the baseline (5.98), DAC (4.70), and TCPT (4.21). This trend holds on SigLIP, where M-PFT yields the lowest ECE (3.69) compared to the baseline (4.31) and TCPT (4.30) at a comparable accuracy (81.08), proving that semantic anchoring suppresses novel-class overconfidence without sacrificing generalization.

\begin{table*}[h!]
\vspace{-15pt}
\centering
\resizebox{\textwidth}{!}{%
\begin{tabular}{@{}l|c|c|c|c|ccccccccc|>{\columncolor{pink!20}}r@{}}
\toprule
\multirow{2}{*}{\textbf{Class}} & \multirow{2}{*}{\textbf{Zero-Shot}} & \multirow{2}{*}{\textbf{Calibration}} & \multirow{2}{*}{\textbf{Model}} & \multirow{2}{*}{\textbf{Metric}} & \multirow{2}{*}{\textbf{CoOp}} & \multirow{2}{*}{\textbf{CoCoOp}} & \multirow{2}{*}{\textbf{KgCoOp}} & \multirow{2}{*}{\textbf{MaPLe}} & \multirow{2}{*}{\textbf{ProDA}} & \multirow{2}{*}{\textbf{ProGrad}} & \multirow{2}{*}{\textbf{PromptSRC}} & \multirow{2}{*}{\textbf{HiCroPL}} & \multirow{2}{*}{\textbf{MMRL}} &\\
& & & & & & & & & & & & & & \multirow{-2}{*}{\textit{Average}} \\
\midrule\midrule

\multirow{14}{*}{\textbf{Base}} & \multirow{7}{*}{\shortstack{ 69.21\\ Acc$\uparrow$\\3.79\\ ECE$\downarrow$\\CLIP }} & \multirow{4}{*}{Baseline} & \multirow{2}{*}{CLIP} & Acc. $\uparrow$ & 81.54 & 80.57 & 81.72 & 82.41 & 78.24 & 76.61 & 84.72 & 86.20 & 86.37 & 82.04 \\

& & & & ECE $\downarrow$ & 6.82 & 4.52 & 6.37 & 3.29 & 4.02 & 5.15 & 4.26 & 3.81 & 2.66 & 4.54 \\ 

\cline{4-15}

& & & \multirow{2}{*}{SigLIP} & Acc. $\uparrow$  & 85.28 & 85.39 & 85.96 & 86.11 & 85.82 & 84.07 & 88.61 & 88.90 & 88.51 & 86.52 \\

& & & & ECE $\downarrow$ & 2.54 & 2.51 & 3.33 & 2.56 & 4.20 & 4.12 & 2.11 & 1.82 & 1.60 & 2.75 \\ 

\cline{3-15}

& & \multirow{4}{*}{TCPT} & \multirow{2}{*}{CLIP} & Acc. $\uparrow$ & 83.19 & 80.16 & 81.34 & 82.75 & 78.51 & 76.98 & 85.31 & 86.66 & 86.39 & 82.37 \\

& & & & ECE $\downarrow$ & 2.98 & 4.08 & 5.80 & 2.78 & 3.50 & 4.88 & 3.04 & 3.23 & 3.24 & 3.73 \\

\cline{2-2} \cline{4-15}

& \multirow{7}{*}{\shortstack{ 77.67\\ Acc$\uparrow$\\4.64\\ ECE$\downarrow$\\SigLIP }} & & \multirow{2}{*}{SigLIP} & Acc. $\uparrow$ & 85.48 & 85.23 & 85.97 & 86.19 & 85.85 & 84.56 & 88.88 & 89.35 & 88.60 & 86.68 \\

& & & & ECE $\downarrow$ & 2.79 & 2.65 & 3.49 & 1.67 & 3.28 & 3.62 & 2.25 & 1.97 & 1.99 & 2.63 \\ 

\cline{3-15}

& & \cellcolor{pink!20} & \cellcolor{pink!20} & \cellcolor{pink!20} \textbf{Acc. $\uparrow$} & \cellcolor{pink!20} 82.97 & \cellcolor{pink!20} 79.92 & \cellcolor{pink!20} 81.70 & \cellcolor{pink!20} 82.38 & \cellcolor{pink!20} 80.02 & \cellcolor{pink!20} 77.67 & \cellcolor{pink!20} 84.91 & \cellcolor{pink!20} 86.25 & \cellcolor{pink!20} 86.42 & 82.47 \\

& & \cellcolor{pink!20} & \cellcolor{pink!20} \multirow{-2}{*}{\textbf{CLIP}} & \cellcolor{pink!20} \textbf{ECE $\downarrow$} & \cellcolor{pink!20} 2.94 & \cellcolor{pink!20} 4.16 & \cellcolor{pink!20} 5.94 & \cellcolor{pink!20} 2.80 & \cellcolor{pink!20} 4.50 & \cellcolor{pink!20} 5.32 & \cellcolor{pink!20} 3.30 & \cellcolor{pink!20} 3.16 & \cellcolor{pink!20} 2.50 & 3.85 \\

\cline{4-15}

& & \cellcolor{pink!20} & \cellcolor{pink!20} & \cellcolor{pink!20} \textbf{Acc. $\uparrow$} & \cellcolor{pink!20} 85.23 & \cellcolor{pink!20} 85.04 & \cellcolor{pink!20} 85.79 & \cellcolor{pink!20} 86.31 & \cellcolor{pink!20} 85.66 & \cellcolor{pink!20} 83.63 & \cellcolor{pink!20} 88.43 & \cellcolor{pink!20} 88.87 & \cellcolor{pink!20} 88.46 & 86.38 \\ 

& & \cellcolor{pink!20} \multirow{-4}{*}{\textbf{M-PFT (Ours)}} & \cellcolor{pink!20} \multirow{-2}{*}{\textbf{SigLIP}} & \cellcolor{pink!20} \textbf{ECE $\downarrow$} & \cellcolor{pink!20} 2.79 & \cellcolor{pink!20} 2.36 & \cellcolor{pink!20} 3.24 & \cellcolor{pink!20} 1.95 & \cellcolor{pink!20} 3.56 & \cellcolor{pink!20} 3.51 & \cellcolor{pink!20} 1.99 & \cellcolor{pink!20} 1.90 & \cellcolor{pink!20} 1.60 & 2.54 \\

\midrule

\multirow{15}{*}{\textbf{Novel}} & \multirow{9}{*}{ \shortstack{ 74.92\\ Acc$\uparrow$\\[1mm] 4.65\\ ECE$\downarrow$\\CLIP }} & \multirow{4}{*}{Baseline} & \multirow{2}{*}{CLIP} & Acc. $\uparrow$ & 69.25 & 73.44 & 74.07 & 75.64 & 70.22 & 74.03 & 75.80 & 77.52 & 76.88 & 74.09 \\

& & & & ECE $\downarrow$ & 12.63 & 5.55 & 3.90 & 6.18 & 9.03 & 4.52 & 4.57 & 3.69 & 3.72 & 5.98 \\

\cline{4-15}

& & & \multirow{2}{*}{SigLIP} & Acc. $\uparrow$  & 76.16 & 79.36 & 79.53 & 80.28 & 78.36 & 78.24 & 82.66 & 83.74 & 82.32 & 80.07  \\

& & & & ECE $\downarrow$ & 6.94 & 5.57 & 3.48 & 4.66 & 4.63 & 3.42 & 3.58 & 2.77 & 3.73 & 4.31 \\ 

\cline{3-15}

& & DAC & CLIP & ECE $\downarrow$ & 7.13 & 5.15 & 4.57 & 4.86 & 4.15 & 3.91 & 3.81 & 3.75 & 4.95 & 4.70 \\

\cline{3-15}

& & \multirow{4}{*}{TCPT} & \multirow{2}{*}{CLIP} & Acc. $\uparrow$ & 69.50 & 73.50 & 73.80 & 75.62 & 71.38 & 74.25 & 75.63 & 77.59 & 75.43 & 74.08 \\

\cline{2-2}

& \multirow{7}{*}{\shortstack{ 81.67\\ Acc$\uparrow$\\3.22\\ ECE$\downarrow$\\SigLIP }} & & & ECE $\downarrow$ & 5.06 & 3.84 & 3.48 & 4.48 & 3.42 & 3.70 & 3.80 & 4.03 & 6.10 & 4.21 \\

\cline{4-15}

& & & \multirow{2}{*}{SigLIP} & Acc. $\uparrow$ & 75.82 & 79.57 & 78.84 & 82.27 & 79.47 & 79.61 & 82.81 & 84.00 & 82.85 & 80.58 \\ 

& & & & ECE $\downarrow$  & 7.35 & 4.71 & 4.47 & 3.33 & 4.30 & 3.48 & 4.46 & 2.99 & 3.64 & 4.30 \\ 

\cline{3-15}

& & \cellcolor{pink!20} & \cellcolor{pink!20} & \cellcolor{pink!20} \textbf{Acc. $\uparrow$} & \cellcolor{pink!20} 72.68 & \cellcolor{pink!20} 74.61 & \cellcolor{pink!20} 74.10 & \cellcolor{pink!20} 76.10 & \cellcolor{pink!20} 75.74 & \cellcolor{pink!20} 73.89 & \cellcolor{pink!20} 76.18 & \cellcolor{pink!20} 77.81 & \cellcolor{pink!20} 76.81 & 75.32 \\

& & \cellcolor{pink!20} & \cellcolor{pink!20} \multirow{-2}{*}{\textbf{CLIP}} & \cellcolor{pink!20} \textbf{ECE $\downarrow$} & \cellcolor{pink!20} 4.75 & \cellcolor{pink!20} 3.79 & \cellcolor{pink!20} 3.72 & \cellcolor{pink!20} 4.08 & \cellcolor{pink!20} 3.99 & \cellcolor{pink!20} 3.51 & \cellcolor{pink!20} 3.74 & \cellcolor{pink!20} 3.71 & \cellcolor{pink!20} 3.47 & 3.86 \\

\cline{4-15}

& & \cellcolor{pink!20} & \cellcolor{pink!20} & \cellcolor{pink!20} \textbf{Acc. $\uparrow$} & \cellcolor{pink!20} 78.61 & \cellcolor{pink!20} 81.18 & \cellcolor{pink!20} 79.59 & \cellcolor{pink!20} 82.29 & \cellcolor{pink!20} 78.70 & \cellcolor{pink!20} 79.74 & \cellcolor{pink!20} 83.64 & \cellcolor{pink!20} 83.80 & \cellcolor{pink!20} 82.16 & 81.08 \\ 

& & \cellcolor{pink!20} \multirow{-4}{*}{\textbf{M-PFT (Ours)}} & \cellcolor{pink!20} \multirow{-2}{*}{\textbf{SigLIP}} & \cellcolor{pink!20} \textbf{ECE $\downarrow$} & \cellcolor{pink!20} 4.87 & \cellcolor{pink!20} 3.48 & \cellcolor{pink!20} 3.81 & \cellcolor{pink!20} 3.49 & \cellcolor{pink!20} 4.18 & \cellcolor{pink!20} 3.67 & \cellcolor{pink!20} 3.23 & \cellcolor{pink!20} 2.92 & \cellcolor{pink!20} 3.55 & 3.69 \\

\bottomrule

\end{tabular}%
}
\vspace{-5pt}
\caption{Comparison of calibration performance on base and novel classes averaged across 10 fine-grained classification datasets for top@1 Accuracy (Acc.) and Expected Calibration Error (ECE) with multiple prompt-tuning methods across various calibration baselines, where higher Acc. and lower ECE indicate better performance and calibration, respectively}
\label{tab:results_avg}
\vspace{-10pt}
\end{table*}

\textbf{Calibration performance under natural distribution shifts.} Tab.~\ref{tab:shift} shows average accuracy and ECE under natural distribution shifts across 5 ImageNet variants with MaPLe~\cite{khattak2023maple}. M-PFT achieves the lowest average ECE of \(2.03\) and \(1.95\) and higher average accuracy of \(71.20\) and \(69.92\) for base and novel classes, respectively. This shows that M-PFT outperforms Zero-Shot, Baseline, and TCPT, highlighting its robustness in generalizing to out-of-distribution scenarios.

\begin{table}[h!]
\vspace{-5pt}
\centering
\resizebox{0.75\linewidth}{!}{%
\begin{tabular}{@{}l|c|ccc|>{\columncolor{pink!20}}r@{}}
\toprule
\multirow{2}{*}{\textbf{Class}} & \multirow{2}{*}{\textbf{Metric}} & \multirow{2}{*}{\textbf{Zero-Shot}} & \multirow{2}{*}{\textbf{Baseline}} & \multirow{2}{*}{\textbf{TCPT}} & \textbf{M-PFT} \\ 
& & & & & \textbf{(Ours)} \\
\midrule\midrule

\multirow{2}{*}{\textbf{Base}} & Acc. $\uparrow$ & 60.52 & 70.19 & 70.04 & 71.20 \\

& ECE $\downarrow$ & 3.92 & 3.00 & 2.73 & 2.03 \\

\midrule

\multirow{2}{*}{\textbf{Novel}} &  Acc. $\uparrow$ & 60.52 & 68.95 & 68.77 & 69.92 \\

& ECE $\downarrow$ & 4.03 & 2.90 & 2.55 & 1.95 \\

\bottomrule

\end{tabular}%
}
\vspace{-5pt}
\caption{Calibration performance under natural distribution shifts.}
\label{tab:shift}
\vspace{-10pt}
\end{table}

\textbf{Calibration performance with different backbones.} Tab.~\ref{tab:backbone} shows that M-PFT achieves better novel-class calibration compared to the Baseline and TCPT with CoOp, achieving the lowest average ECE of \(4.96\) for CLIP RN-50 and \(4.70\) for CLIP ViT-B/32. For base classes, M-PFT remains competitive, achieving an ECE of \(3.78\) and \(2.98\) for RN-50 and ViT-B/32. These findings suggest that M-PFT provides robust calibration across different backbones. 

\subsection{Ablation Studies}
\label{sec:ablation}

\textbf{Loss component analyses.}
Tab.~\ref{tab:gradient_avg} analyzes the contribution of each loss component to calibration across \(10\) datasets with MaPLe. All variants are built on a cross-entropy baseline. Adding \(\mathcal{L}_{\text{min-max}}\) alone reduces base-class ECE from \(3.29\) to \(2.49\) while maintaining accuracy at \(81.74\), confirming that maximizing angular distance improves class discriminability. However, novel-class ECE remains high, ranging from \(6.18\) to \(6.80\), showing that promoting class discriminability without semantic constraints can amplify overconfidence on unseen classes. \begin{table}[t!]
\centering
\resizebox{0.7\linewidth}{!}{%
\begin{tabular}{@{}l|c|c|cc|>{\columncolor{pink!20}}r@{}}
\toprule
\multirow{2}{*}{\textbf{Class}} & \multirow{2}{*}{\textbf{Backbone}} & \multirow{2}{*}{\textbf{Metric}} & \multirow{2}{*}{\textbf{Baseline}} & \multirow{2}{*}{\textbf{TCPT}} & \textbf{M-PFT} \\ 
& & & & & \textbf{(Ours)} \\
\midrule\midrule

\multirow{4}{*}{\textbf{Base}} & \multirow{2}{*}{ResNet-50} & Acc. $\uparrow$ & 77.66 & 78.57 & 78.14 \\

&  & ECE $\downarrow$ & 4.04 & 3.46 & 3.78 \\ 

\cline{2-6} 
& \multirow{2}{*}{ViT-B/32} & Acc. $\uparrow$ & 80.56 & 80.90 & 80.16 \\

&  & ECE $\downarrow$ & 3.15 & 2.87 & 2.98 \\ 

\midrule

\multirow{4}{*}{\textbf{Novel}} & \multirow{2}{*}{ResNet-50} & Acc. $\uparrow$ & 59.57 & 60.68 & 65.66 \\

&  & ECE $\downarrow$ & 9.38 & 5.46 & 4.96 \\ 

\cline{2-6} 

& \multirow{2}{*}{ViT-B/32} & Acc. $\uparrow$ & 65.09 & 65.63 & 69.45 \\

&  & ECE $\downarrow$ & 10.13 & 5.82 & 4.70 \\ 

\bottomrule
\end{tabular}%
}
\vspace{-5pt}
\caption{Calibration performance with different backbones.}
\label{tab:backbone}
\vspace{-15pt}
\end{table}
 Adding the \(\mathcal{L}_{\text{anchor}}\) only variant preserves semantic geometry, lowering novel ECE from \(6.18\) to \(4.28\), but base ECE remains high at 3.25 due to the collapsed embedding. Adding \(\mathcal{L}_{\text{anchor}}\) to \(\mathcal{L}_{\text{min-max}}\) improves novel-class calibration. Without gradient blocking, novel ECE reduces from \(6.80\) to \(4.11\) with accuracy of \(75.88\), but base ECE remains high, from \(2.49\) to \(3.12\), because without blocking, zero-shot embeddings are pulled toward tunable embeddings, leading to clustering and reduced class discriminability \(\theta_{jk}\downarrow\). With gradient blocking, the full M-PFT objective further reduces novel ECE from \(4.11\) to \(4.08\) and base ECE from \(3.12\) to \(2.80\), yielding a Pareto-optimal base-to-novel calibration trade-off across all datasets. Since the variant without gradient blocking achieves lower base and novel class ECE on some datasets like SUN397, Aircraft, and Caltech101 (see suppl. Tab.~\ref{tab:gradient}), confirming that gradient blocking is necessary but not always.
Overall, \(\mathcal{L}_{\text{min-max}}\) improves base-class calibration, \(\mathcal{L}_{\text{anchor}}\) reduces novel-class overconfidence, and gradient blocking enables both objectives to operate synergistically, collectively constituting the full M-PFT objective. 

\begin{table}[t!]
\vspace{-5pt}
\centering
\resizebox{0.75\linewidth}{!}{%
\begin{tabular}{@{}l|c|c|c|r@{}}
\toprule
\multirow{2}{*}{\textbf{Loss
Component}}& \multicolumn{2}{c|}{\textbf{Base}} & \multicolumn{2}{c}{\textbf{Novel}} \\

\cline{2-5}

& Acc. $\uparrow$ & ECE $\downarrow$ & Acc. $\uparrow$ & ECE $\downarrow$ \\
\midrule\midrule

\(\mathcal{L}_{\text{CE}}\) &  82.41 & 3.29 & 75.64 & 6.18 \\

\midrule

\(\mathcal{L}_{\text{CE}} + \mathcal{L}_{\text{min-max}}\) & 81.74 & 2.49 & 72.06 & 6.80 \\

\midrule

\(\mathcal{L}_{\text{CE}} + \mathcal{L}_{\text{anchor}}\) & 82.64 & 3.25 & 75.36 & 4.28 \\

\midrule

\(\mathcal{L}_{\text{CE}} + \mathcal{L}_{\text{min-max}} + \mathcal{L}_{\text{anchor}}\) & \multirow{2}{*}{82.43} & \multirow{2}{*}{3.12} & \multirow{2}{*}{75.88} & \multirow{2}{*}{4.11} \\

(w/o gradient blocking) & & & & \\

\midrule

\rowcolor{pink!20} \textbf{\(\mathcal{L}_{\text{CE}} + \mathcal{L}_{\text{min-max}} + \mathcal{L}_{\text{anchor}}\)} & & & & \\

\rowcolor{pink!20} \textbf{(w/ gradient blocking)} & \multirow{-2}{*}{82.38} & \multirow{-2}{*}{2.80} & \multirow{-2}{*}{76.10} & \multirow{-2}{*}{4.08} \\

\bottomrule

\end{tabular}%
}
\vspace{-5pt}
\caption{Loss component analyses on the effect of min-max regularization \(\mathcal{L}_{\text{min-max}}\) and anchoring regularization \(\mathcal{L}_{\text{anchor}}\) with (w/) and without (w/o) gradient blocking
on both base and novel classes.}
\label{tab:gradient_avg}
\vspace{-15pt}
\end{table}

\begin{figure}[t!]
\vspace{-10pt}
\centering
\includegraphics[width=0.75\linewidth]{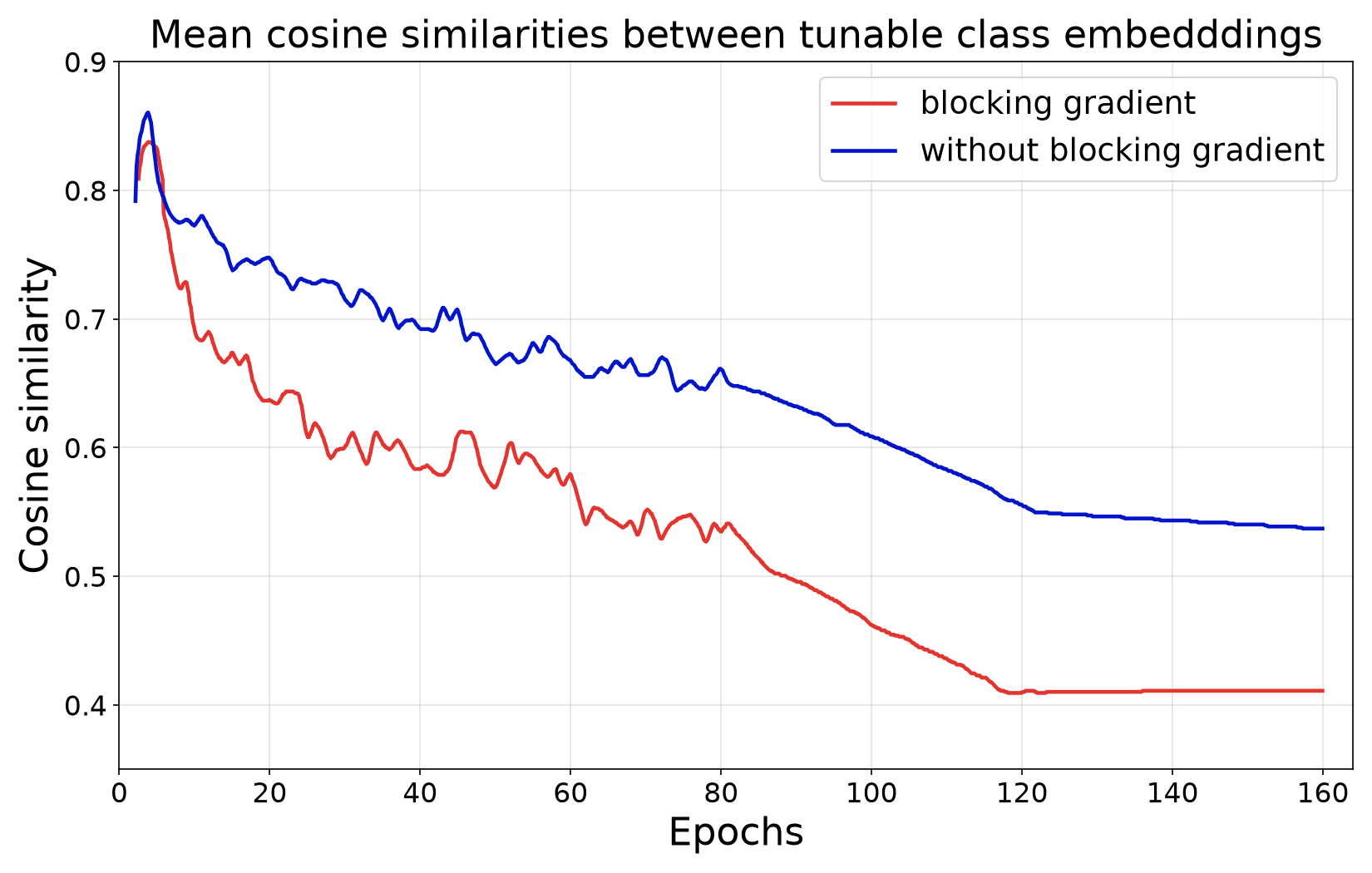}
\vspace{-10pt}
\caption{Cosine similarity comparision between tunable class embeddings. With gradient blocking w.r.t. zero-shot embeddings, cosine similarity is lower, corresponding to increased \(\theta_{jk}\uparrow\), showing higher class discriminability.}
\label{fig:gradient}
\vspace{-15pt}
\end{figure}

\textbf{Gradient blocking graph.} We use the cosine similarity between tunable class embeddings to measure class discriminability. Fig.~\ref{fig:gradient} compares cosine similarity with and without blocking the gradient of the anchoring regularization term w.r.t. the zero-shot embedding. The mean cosine similarity between tunable class embeddings is lower with gradient blocking compared to without gradient blocking. This indicates that gradient blocking has larger class discriminability \(\theta_{jk}\), consistent with the analysis in Sec.~\ref{sec:method}. As shown in Tab.~\ref{tab:gradient_avg}, gradient blocking also achieves a Pareto-optimal base-to-novel calibration with lower ECE on both base and novel classes. However, the counterpart increases the ECE, particularly on base classes. These results show the importance of blocking the gradient w.r.t. the zero-shot embedding.

\textbf{Anchor-based methods.} Prior anchor-based methods like KgCoOp, ProGrad, and PromptSRC apply distance, alignment, or consistency to improve generalization to novel classes without connecting angular geometry to the base-to-novel calibration trade-off. We link angular geometry: inter-class embedding collapse \(\theta_{jk} \rightarrow 0\) shrinks top-1 and runner-up class margins (\(p_j \rightarrow p_k\)), capping base confidence, while semantic drift \(\theta_j \uparrow\) inflates margins to spike novel confidence. As shown in Tab. 5, on average across these three methods, M-PFT reduces CLIP base ECE from 5.26 to 4.85 at 81.43 accuracy and novel ECE from 4.33 to 3.66 at 74.72 accuracy. For SigLIP, it achieves a base ECE of 2.91 at 85.95 accuracy and a novel ECE of 3.57 at 80.99 accuracy.

\begin{table}[h!]
\vspace{-5pt}
\centering
\centering
\resizebox{\linewidth}{!}{%
\begin{tabular}{@{}l|c|c|>{\columncolor{pink!20}}c|cc|>{\columncolor{pink!20}}c|>{\columncolor{pink!20}}c@{}}
\toprule
\multirow{3}{*}{\textbf{Class}} & \multirow{3}{*}{\textbf{Metric}} & \multicolumn{2}{c|}{\textbf{Standard deviations (Std.))}} & \multicolumn{4}{c}{\textbf{Anchor-based Methods}} \\ \cline{3-8} 
 &  & \multirow{2}{*}{\textbf{Baseline}} & \textbf{M-PFT} & \multicolumn{2}{c}{\textbf{Baseline}} & \multicolumn{2}{|>{\columncolor{pink!20}}c}{\textbf{M-PFT (Ours)}} \\
 \cline{5-8}
 &  &  & \textbf{(Ours)} & \textbf{CLIP} & \textbf{SigLIP} & \textbf{CLIP} & \textbf{SigLIP} \\ \midrule
\multirow{2}{*}{\textbf{Base}} & Acc. $\uparrow$ & $\pm$ 0.36 & $\pm$ 0.35 & 81.02 & 86.21 & 81.43 & 85.95 \\
 & ECE $\downarrow$ & $\pm$ 0.29 & $\pm$ 0.26 & 5.26 & 3.17 & 4.85 & 2.91 \\ \midrule
\multirow{2}{*}{\textbf{Novel}} & Acc. $\uparrow$ & $\pm$ 0.51 & $\pm$ 0.52 & 74.63 & 80.14 & 74.72 & 80.99 \\
 & ECE $\downarrow$ & $\pm$ 0.47 & $\pm$ 0.42 & 4.33 & 3.49 & 3.66 & 3.57 \\ \bottomrule
\end{tabular}%
}
\vspace{-5pt}
\caption{Standard deviation and Anchor-based method analyses.}
\label{tab:anchor}
\vspace{-10pt}
\end{table}

\textbf{Computational overhead.} We evaluate M-PFT's computational and memory overhead via per-batch calculation time, GPU memory usage, and ECE. M-PFT marginally increases calculation time, leaves memory usage essentially unchanged, and reduces ECE, demonstrating negligible overhead (see suppl. for details).

\textbf{Impact of hyperparameters.} We investigate the impact of different regularization hyperparameters on M-PFT calibration. Overly large hyperparameters may slightly improve or degrade performance, as too strong regularization can prevent training with standard cross-entropy loss. Hence, we set \(\lambda_1=0.4\) and \(\lambda_2=1.4\) via hyperparameter search on 3 of 15 datasets and held fixed across all remaining datasets in the following experiments. M-PFT shows strong cross-dataset transferability, is not highly sensitive to either hyperparameter, and shows a stable ECE for \(\lambda_1 \in \left[0.4,~1.0\right]\) and \(\lambda_2 \in \left[0.5,~1.4\right]\), showing robustness without dataset-specific tuning. (see suppl. for details) 

\textbf{Standard deviation (Std.) analyses.} Tab.~\ref{tab:anchor} reports average standard deviations of accuracy and ECE across 10 datasets for 3 seed runs with MMRL. M-PFT yields lower average ECE variance than the baseline on base \(\pm0.29 \to \pm0.26\) and novel \(\pm0.47 \to \pm0.42\) classes, with a competitive accuracy variance of \(\pm0.35\). This demonstrates superior stability w.r.t. random seed initialization.

\begin{figure}[h!]
\vspace{-5pt}
\centering
\begin{subfigure}{0.49\linewidth}
\centering    
\includegraphics[width=\linewidth]{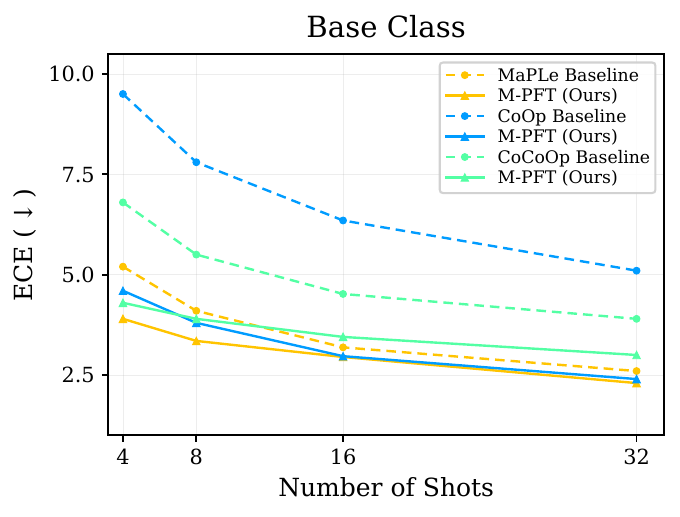}
\end{subfigure}
\begin{subfigure}{0.49\linewidth}
\centering
\includegraphics[width=\linewidth]{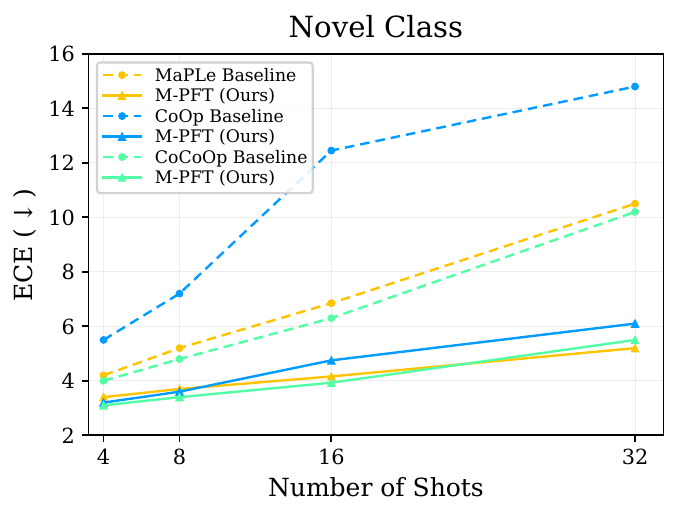}
\end{subfigure}
\vspace{-10pt}
\caption{Varying shots.}
\label{fig:shots}
\vspace{-10pt}
\end{figure}

\textbf{Varying shots.} Fig.~\ref{fig:shots} shows calibration across different \(4\), \(8\), \(16\), and \(32\) numbers of shots with MaPLe, CoOp, and CoCoOp. Baselines exhibit higher ECE with fewer shots on base classes and increasing ECE at higher shots on novel classes. In contrast, M-PFT maintains lower and stable ECE across \(4\), \(8\), \(16\), and \(32\)-shot settings and backbones, showing superior calibration even in extreme few-shot scenarios. However, calibration gains may diminish under extreme domain gaps and extremely few-shot 1-shot scenarios.

\begin{figure}[h!]
\vspace{-5pt}
\centering
\begin{subfigure}{0.49\linewidth}
\centering    
\includegraphics[width=\linewidth]{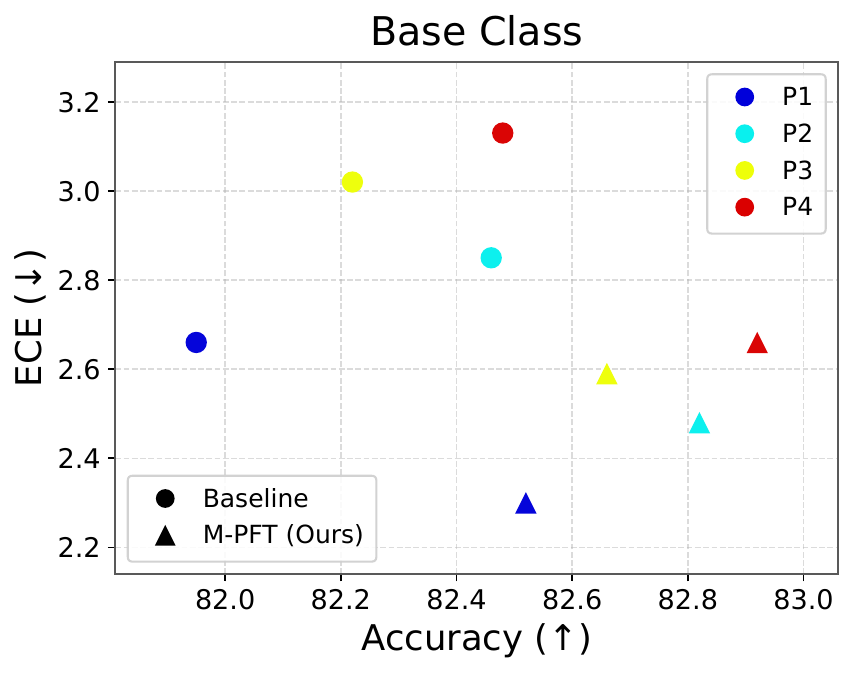}
\end{subfigure}
\begin{subfigure}{0.49\linewidth}
\centering
\includegraphics[width=\linewidth]{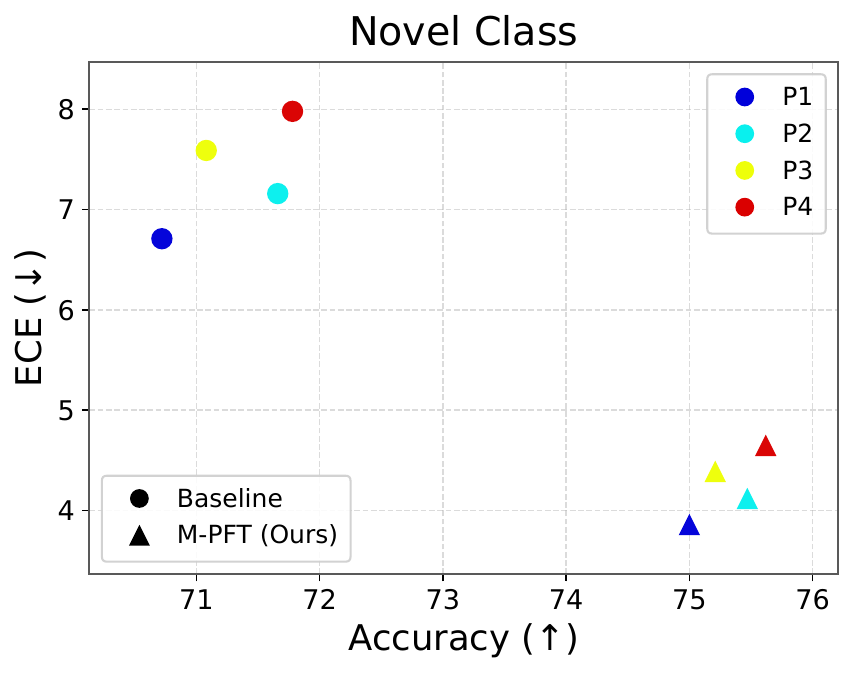}
\end{subfigure}
\vspace{-10pt}
\caption{Different prompt initialization.}
\label{fig:prompt}
\vspace{-10pt}
\end{figure}

\textbf{Different prompt initializations.} We evaluate sensitivity across four hand-crafted templates (P1–P4) on 10 datasets with MaPLe (Fig.~\ref{fig:prompt}). M-PFT consistently yields lower ECE and higher accuracy with minimal fluctuation compared to the sensitive baseline. This demonstrates robustness rather than template overfitting (see suppl. for details).

\section{Conclusion}

In this paper, we propose M-PFT, a novel method for calibrated prompt tuning of fine-tuned VLMs that introduces two regularizers: (1) min-max regularization to improve class discriminability and (2) anchoring regularization to preserve the semantic geometry of the pre-trained embedding space. Compared with current methods, M-PFT promotes class discriminability \(\theta_{jk}\uparrow\) while reducing semantic drift \(\theta_j\downarrow\), thereby effectively enlarging the angular margin. This dual design with gradient blocking creates two complementary objectives that act synergistically to exploit the angular discriminative properties of the hyperspherical embedding space without distorting decision boundaries. Extensive experiments across 9 prompt-tuning techniques and 15 datasets demonstrate that M-PFT improves the calibration of prompt-tuned VLMs, particularly on novel classes, without compromising classification accuracy.

\textbf{Acknowledgments.} We acknowledge MBZUAI for providing funding and compute support for this research. The Ministry of Higher Education, Sri Lanka, funded by the World Bank (AHEAD Operation Grant No. 6026-LK/8743-LK), the National Research Council of Sri Lanka (Grant No. 19-080), and Manoj Bandara also partially supported the computational resources.



{
    \small
    \bibliographystyle{ieeenat_fullname}
    \bibliography{main}
}

\clearpage

\appendix

\maketitlesupplementary
\setcounter{page}{1}
\setcounter{section}{0}
\setcounter{table}{0}
\setcounter{figure}{0}
\setcounter{equation}{0}

\section*{Appendix} 

\noindent This supplementary material provides the following:

\begin{itemize}
\item Experimental setup in Sec.~\ref{sec:experimental}
\item Calibration analysis with CLIP and SigLIP backbones for base and novel classes across 10 fine-grained datasets in Sec.~\ref {sec:dataset}.
\item Calibration performance under natural distribution
shifts in Sec.~\ref{sec:shift}.
\item Other calibration metrics, like MCE and ACE in Sec.~\ref{sec:mce_ace}.
\item Loss component analyses in Sec.~\ref{sec:loss}.
\item Impact of hyperparameters in Sec.~\ref{sec:hyperparameter}.
\item Results on various backbones in Sec.~\ref{sec:backbone}.
\item Different prompt initializations in Sec.~\ref{sec:prompt}.
\item Computational overhead in Sec.~\ref{sec:overhead}
\item Standard deviation across random seeds in Sec.~\ref{sec:std}
\end{itemize}

\section{Experimental setup}
\label{sec:experimental}

\textbf{Evaluation setup.} We follow the standard open-vocabulary setting, and the datasets are divided into disjoint base and novel classes. The VLM is prompt-tuned only on base classes in a few-shot setting with a limited number of labeled samples per class, and we report calibration performance on both the base and novel classes.

\textbf{Datasets.} We evaluate the proposed approach along with comparison calibration methods on 15 datasets spanning various image classification tasks. These tasks include object recognition on ImageNet~\cite{deng2009imagenet} and Caltech\(101\)~\cite{fei2004learning} datasets, fine-grained object recognition such as Oxford Pets~\cite{parkhi2012cats}, Food\(101\)~\cite{bossard2014food}, Stanford Cars~\cite{krause20133d}, Oxford Flowers\(102\)~\cite{nilsback2008automated}, and FGVCAircraft~\cite{maji2013fine} datasets. Additionally, we performed a remote sensing recognition task with the EuroSAT~\cite{helber2018introducing} dataset, a texture recognition task with the DTD~\cite{cimpoi2014describing} dataset, an action recognition task with the UCF\(101\)~\cite{soomro2012ucf101} dataset, and a large-scale scene understanding task with the SUN\(397\) dataset. And we also provide results on out-of-distribution tasks, including ImageNet-V2~\cite{recht2019imagenet} (natural images), ImageNet-A~\cite{hendrycks2021natural} (natural adversarial examples), ImageNet-R~\cite{hendrycks2021many} (artistic renditions), and ImageNet-S~\cite{wang2019learning} (black and white sketches) datasets. Following the base-to-new generalization protocol introduced by CoCoOp~\cite{zhou2022conditional}, downstream datasets are partitioned into disjoint base and novel class subsets. As detailed in Tab.~\ref{tab:splits}, each dataset is divided into training and test splits. The fine-tuned CLIP model is prompt-tuned solely on the base classes. Then the model is tested on both class subsets and unseen categories.

\begin{table}[h!]
\centering
\resizebox{\linewidth}{!}{%
\begin{tabular}{@{}l|c|c|c|r@{}}
\toprule
\multirow{2}{*}{\textbf{Dataset}} & \multicolumn{3}{c|}{\textbf{Class Counts}} & \textbf{Train Samples} \\

\cline{2-5} 

& \textbf{Base} & \textbf{Novel} & \textbf{Total} & \textbf{Base (16-shot)} \\
\midrule

\multicolumn{4}{@{}l@{}}{\textit{Generic Objects}} \\

~~Caltech101 & 50 & 50 & 100 & 800 \\

~~ImageNet & 500 & 500 & 1,000 & 8,000 \\

\midrule

\multicolumn{4}{@{}l@{}}{\textit{Fine-Grained}} \\

~~OxfordPets & 19 & 18 & 37 & 304 \\

~~StanfordCars & 98 & 98 & 196 & 1,568 \\

~~Flowers102 & 51 & 51 & 102 & 816 \\

~~Food101 & 51 & 50 & 101 & 816 \\

~~FGVCAircraft & 50 & 50 & 100 & 800 \\

\midrule

\multicolumn{4}{@{}l@{}}{\textit{Specialized Contexts}} \\

~~SUN397 & 199 & 198 & 397 & 3,184 \\

~~DTD & 24 & 23 & 47 & 384 \\

~~EuroSAT & 5 & 5 & 10 & 80 \\

~~UCF101 & 51 & 50 & 101 & 816 \\

\bottomrule

\end{tabular}%
}
\caption{The split strategy of downstream datasets categorized by domain. Prompt tuning utilizes few-shot \((16)\) training samples only from the base classes ($\mathcal{C}_{\text{base}}$), while evaluated across separate base and novel ($\mathcal{C}_{\text{novel}}$) test splits.}
\label{tab:splits}
\end{table}

\textbf{Compared methods.} We benchmark M-PFT against a diverse set of representative prompt tuning methods: CoOp~\cite{zhou2022learning}, CoCoOp~\cite{zhou2022conditional}, KgCoOp~\cite{yao2023visual}, MaPLe~\cite{khattak2023maple}, ProDA~\cite{lu2022prompt}, ProGrad~\cite{zhu2023prompt}, PromptSRC~\cite{khattak2023self}, HiCroPL~\cite{zheng2025hierarchical}, and MMRL~\cite{guo2025mmrl}, and calibration baselines. We list the general hyperparameters in Tab~\ref{tab:vlm_hyperparameters}. Here, we briefly introduce the corresponding exclusive hyperparameters of each VLM tuning method. All the methods are adopted from their official implementation. For CoOp and CoCoOp, they do not contain other hyperparameters. For ProDA, we set \(\lambda = 0.1\). For KgCoOp, we set \(\lambda = 8.0\). For MaPLe, we set prompt depth \(J\) to \(0\) and the language and vision prompt lengths to \(2\). For ProGrad, we set \(\lambda = 0.8\). For PromptSRC, we set deep prompting with \(V = T = 4\). \(\lambda_1 = 10\) and \(\lambda_2 = 25\) are used in the weight loss. For textual diversity, we use a total of \(N = 60\) standard prompt templates. Since methods like CoOp only consider the text modality and exhibit low accuracy, making their calibration significance minimal, we focused on testing and calibrating their optimized versions, CoCoOp and KgCoOp. For calibration, we select representative baselines: classical baselines include a post-hoc approach, DAC~\cite{wang2024open} for adapting CLIP to novel classes, and a training-time regularization method, TCPT~\cite{sharifdeen2026towards}.

\begin{table}[h!]
\centering
\begin{tabular}{@{}lcccr@{}}
\toprule
\textbf{Methods} & \textbf{BS} & \textbf{LR} & \textbf{CTX} \\ \midrule
CoOp & 32 & 0.002 & 16 \\
CoCoOp & 1 & 0.002 & 4 \\
ProDA & 4 & 0.001 & 16 \\
KgCoOp & 32 & 0.002 & 16 \\
MaPLe & 4 & 0.0026 & 2 \\
ProGrad & 32 & 0.002 & 16 \\
PromptSRC & 4 & 0.0025 & 4 \\ 
\bottomrule
\end{tabular}
\caption{Hyperparameters for prompt tuning methods. ``BS'' denotes the batch size. ``LR'' denotes the learning rate. ``CTX'' is the context length of the learnable prompt.}
\label{tab:vlm_hyperparameters}
\end{table}

\textbf{Implementation details.} For the main experiments, we use CLIP~\cite{deng2009imagenet} and SigLIP ViT-B/\(16\)~\cite{zhai2023sigmoid} as the pre-trained VLM throughout and report results averaged over \(3\) seed runs. For all compared methods, we use the corresponding official implementations with the specified hyperparameter configurations. Prompt-tuning is conducted in a few-shot manner with \(16\) samples per class, using a learning rate of \(0.005\) and a batch size of \(8\). All experiments employ CLIP with a ViT-B/16~\cite{radford2021learning} as the pre-trained vision-language backbone. 
For each baseline method, we adopt the corresponding official implementation and follow the hyperparameter configurations recommended in the original publications to ensure a fair comparison. All experiments are conducted using PyTorch~\cite{paszke2019pytorch} on a single NVIDIA RTX A\(6000\) GPU with \(48\)GB of memory. 

\textbf{Evaluation metrics.} We use top-1 accuracy (Acc.) for classification performance and Expected Calibration Error (ECE)~\cite{naeini2015obtaining} for open-vocabulary confidence calibration in our evaluation, which is a widely used metric. We also use Adaptive Calibration Error (ACE)~\cite{nixon2019measuring} and Maximum Calibration Error (MCE)~\cite{naeini2015obtaining}, results carried in the supplementary material. All calibration errors are given by $\times10^{-2}$.

\section{Calibration analysis across 10 fine-grained datasets}
\label{sec:dataset}

We evaluate the calibration performance of M-PFT across 10 fine-grained classification tasks, comparing it with \(9\) representative prompt-tuning methods: CoOp~\cite{zhou2022learning}, CoCoOp~\cite{zhou2022conditional}, KgCoOp~\cite{yao2023visual}, MaPLe~\cite{khattak2023maple}, ProDA~\cite{lu2022prompt}, ProGrad~\cite{zhu2023prompt}, PromptSRC~\cite{khattak2023self}, HiCroPL~\cite{zheng2025hierarchical}, and MMRL~\cite{guo2025mmrl}. To assess whether these gains generalize across vision-language backbones, we report results using both \textbf{CLIP}~\cite{radford2021learning} and \textbf{SigLIP}~\cite{zhai2023sigmoid} as the underlying backbone. While the main paper summaries focus on aggregate metrics, we provide comprehensive dataset-wise results in Tab.~\ref{tab:base} and Tab.~\ref{tab:base_siglip} for base classes, while Tab.~\ref{tab:novel} and Tab.~\ref{tab:novel_siglip} for novel classes to demonstrate the broad applicability and efficacy of M-PFT. All reported results are averaged across three random seeds.

\textbf{Calibration performance on base classes (CLIP).} As shown in Tab.~\ref{tab:base}, M-PFT consistently achieves superior calibration performance (lower ECE) while maintaining competitive accuracy across all evaluated prompt learning methods and datasets. For CoOp, M-PFT reduces the average ECE from \(6.82\) (Baseline) to \(2.94\), surpassing TCPT's \(2.98\) while improving accuracy from \(81.54\) to \(82.97\). In methods like CoCoOp and KgCoOp, M-PFT remains competitive with TCPT while consistently outperforming the Baseline, and for MaPLe, it achieves an average ECE of \(2.80\) without sacrificing accuracy \(82.38\). M-PFT also shows strong cross-method compatibility, achieving the best ECE for HiCroPL \(3.16\) and the most significant aggregate gain with MMRL, lowering the average ECE to \(2.50\) compared to \(3.24\) for TCPT. At the dataset level, on Flowers102 within PromptSRC, M-PFT achieves an ECE of \(3.10\) against the baseline's \(5.15\), and on OxfordPets within MMRL, it reaches an accuracy of \(95.82\), confirming that calibration gains do not come at the cost of accuracy. These results validate the generalizability of M-PFT in mitigating base-class underconfidence across diverse prompt-tuning techniques and datasets.

\noindent\textbf{Calibration performance on novel classes (CLIP).}
We further evaluate the robustness and efficacy of M-PFT on novel classes, summarized in Tab.~\ref{tab:novel}. Consistent with the trends observed on base classes, M-PFT maintains a superior balance between accuracy and ECE on unseen categories. For CoOp, M-PFT substantially reduces the average ECE from \(12.63\) (Baseline) to \(4.75\), surpassing both DAC (\(7.13\)) and TCPT (\(5.06\)), with accuracy improving from \(69.25\) to \(72.68\). For MaPLe, M-PFT yields a marked improvement on EuroSAT, lowering ECE from \(11.68\) (Baseline) to \(5.23\) while nearly matching the Baseline's accuracy. While ProDA achieves a notable accuracy gain on EuroSAT (\(51.86\) to \(74.64\)), its overall average ECE (\(3.99\)) trails several other methods, with MMRL achieving the lowest average ECE among all settings at \(3.47\). HiCroPL attains the best overall average accuracy (\(77.81\)) among the compared methods, closely followed by PromptSRC at \(76.18\). Crucially, these calibration gains do not come at the expense of discriminative strength, as M-PFT maintains high accuracy levels such as \(97.84\) on OxfordPets within ProGrad and \(93.52\) on Caltech101 within ProDA. These results confirm that M-PFT remains reliable and well-calibrated even when encountering novel, out-of-distribution classes.

\textbf{Calibration performance on base classes (SigLIP).} To assess whether the dataset-wise calibration gains of M-PFT extend beyond the CLIP backbone, we report the full dataset-wise breakdown using SigLIP in Tab.~\ref{tab:base_siglip}. M-PFT achieves lower or comparable ECE relative to both the Baseline and TCPT across most prompt-tuning methods, while accuracy remains largely on par with both. For CoOp, M-PFT's average ECE rises slightly from 2.54 (Baseline) to 2.79, though a dataset-wise improvement is still visible on Caltech101 (1.25 to 1.01). For KgCoOp, M-PFT reduces ECE on Aircraft from 5.84 to 4.46 while accuracy is essentially preserved. The most pronounced aggregate improvement is observed with MaPLe, where M-PFT lowers the average ECE from 2.56 (Baseline) to 1.95 while simultaneously achieving the highest accuracy among the three settings at 86.31. These results confirm that M-PFT's class-discriminability objective continues to mitigate base-class underconfidence across most prompt-tuning methods under a different vision-language backbone, such as SigLIP.

\textbf{Calibration performance on novel classes (SigLIP).}
We further evaluate the generalization capability of M-PFT to novel classes under the SigLIP backbone, with the complete dataset-wise results presented in Tab.~\ref{tab:novel_siglip}. Consistent with the trends observed on the CLIP backbone, M-PFT achieves the lowest or near-lowest average ECE across nearly all evaluated methods, while improving or maintaining accuracy relative to the Baseline, DAC, and TCPT. For CoOp, M-PFT substantially reduces the average ECE from 6.94 (Baseline) to 4.87, surpassing TCPT 7.35, with accuracy improving from 76.16 to 78.61. For MaPLe, M-PFT yields a marked improvement on EuroSAT, lowering ECE from 14.52 to 9.09 while raising accuracy from 57.51 to 69.65, the largest single-dataset accuracy gain observed in the novel-class setting. PromptSRC achieves the best overall average accuracy of 83.64 among the compared methods, while M-PFT remains competitive with EuroSAT in cross-modal methods such as HiCroPL and MMRL despite the dataset's inherently high variance. These results demonstrate that M-PFT's semantic-anchoring objective continues to suppress novel-class overconfidence at the dataset-wise level under the SigLIP backbone, reinforcing that the dual miscalibration problem and its mitigation generalize beyond CLIP.

\section{Calibration performance under natural distribution shifts}
\label{sec:shift}

We evaluate the robustness of M-PFT under natural distribution shifts across four out-of-distribution (OOD) datasets, representing diverse challenges: natural images (ImageNet-V2)~\cite{recht2019imagenet}, black and white sketches (ImageNet-S)~\cite{wang2019learning}, natural adversarial examples (ImageNet-A),~\cite{hendrycks2021natural}, and artistic renditions (ImageNet-R)~\cite{hendrycks2021many}. As shown in Tab.~\ref{tab:ood}, we compare M-PFT against Zero-Shot, Baseline, DAC, and TCPT across both base and novel classes.

\noindent\textbf{Calibration performance on base classes.} As shown in Tab.~\ref{tab:ood}, M-PFT consistently surpasses both the Baseline and TCPT under natural distribution shifts. M-PFT achieves the best overall calibration with an average ECE of \(2.03\), a significant reduction compared to Zero-Shot \(3.92\), Baseline \(3.00\) and TCPT \(2.73\). These improvements are particularly pronounced on challenging shifts; for instance, M-PFT reduces ECE from \(3.14\) to \(1.52\) on ImageNet-V2, from \(3.94\) to \(2.02\) on ImageNet-S, and from \(3.14\) to \(2.13\) on ImageNet-R. Importantly, these gains are achieved while maintaining competitive accuracy, with M-PFT simultaneously lowering ECE and improving accuracy over the Baseline on ImageNet-S and ImageNet-R, reaching accuracies of \(58.77\) and \(86.06\), respectively.

\noindent\textbf{Calibration performance on novel classes.} The generalization capability of M-PFT is further validated on novel classes. M-PFT reduces ECE from \(3.16\) to \(1.52\) on ImageNet-V2, from \(3.73\) to \(2.05\) on ImageNet-S, and from \(3.14\) to \(2.13\) on ImageNet-R. Consistent with the base-class trends, M-PFT maintains superior performance under distribution shift, achieving the lowest average ECE \(1.95\) and outperforming both the Baseline \(2.90\) and TCPT \(2.55\). The consistent performance across both base and novel classes indicates that M-PFT addresses the fundamental dual miscalibration problem rather than overfitting to dataset-specific artifacts, which is critical for real-world deployment where models must provide reliable confidence estimates even when encountering data that differs from the training distribution.

\section{Other calibration metrics: MCE and ACE}
\label{sec:mce_ace}
To provide a more comprehensive evaluation of model calibration beyond the Expected Calibration Error (ECE) discussed in our main paper, we evaluate M-PFT using two additional metrics: Maximum Calibration Error (MCE) and Adaptive Calibration Error (ACE), both computed with the SigLIP backbone. These metrics offer a deeper look into the reliability of predicted probabilities, with MCE highlighting worst-case scenarios and ACE providing a more robust measure across varying data densities.

As shown in Tab.~\ref{tab:base_mce_ace}, our approach improves calibration performance across most of the \(10\) fine-grained classification datasets compared to prompt-tuning techniques on base classes, when using the SigLIP backbone. Our method significantly refines MaPLe's calibration, with the average MCE dropping from \(0.78\) to \(0.63\) and the average ACE dropping from \(2.47\) to \(1.81\). These shifts indicate a substantial decrease in both average and peak miscalibration. Similar trends were observed with PromptSRC, where M-PFT lowers the average MCE from \(0.68\) to \(0.64\) and the average ACE from \(2.02\) to \(1.82\). For MMRL, M-PFT achieves an average MCE comparable to the baseline \(\left(0.48~\text{vs.}~0.50\right)\) while reducing the average ACE from \(\left(1.54~\text{to}~1.46\right)\). The notable improvements in MCE for MaPLe and PromptSRC are particularly vital. Since MCE identifies the maximum gap between confidence and accuracy, these results suggest that M-PFT effectively mitigates extreme overconfidence, leading to more dependable model outputs even in the most challenging classification cases. Overall, these findings reinforce the versatility and effectiveness of M-PFT in producing well-calibrated models across diverse prompt-learning techniques, even when paired with a different vision-language backbone such as SigLIP.

Analysis of the novel class performance, shown in Tab.~\ref{tab:novel_mce_ace}, largely mirrors the trends observed in the base classes, reinforcing the generalization capability of M-PFT. For MaPLe, M-PFT reduces the average MCE from \(1.58\) to \(1.26\) and the average ACE from \(4.56\) to \(3.44\), again indicating a marked reduction in both peak and average miscalibration. For PromptSRC, M-PFT lowers the average MCE from \(1.05\) to \(0.85\) and the average ACE from \(3.49\) to \(3.11\). For MMRL, M-PFT yields a more modest effect: the average ACE improves slightly from \(3.61\) to \(3.45\), while the average MCE remains comparable to the baseline \(\left(1.22~\text{vs.}~1.27\right)\). This stability across diverse metrics, including ECE, MCE, and ACE, indicates that our method resolves underlying miscalibration rather than merely overfitting to a single error type. By successfully mitigating both worst-case and peak calibration errors across most settings, the proposed method demonstrates the reliability required for deployment in safety-critical applications, regardless of the underlying vision-language backbone.

\section{Loss component analyses}
\label{sec:loss}

\begin{figure}[h!]
\centering

\begin{minipage}{0.47\linewidth} \centering $\mathcal{L}_{\text{CE}}$ \end{minipage}
\begin{minipage}{0.47\linewidth} \centering $\mathcal{L}_{\text{CE}} + \mathcal{L}_{\text{anchor}}$ \end{minipage}

\vspace{1mm}

\rotatebox[origin=l]{90}{\hspace{3.25em} Base class}
\begin{subfigure}{0.47\linewidth}
\includegraphics[width=\linewidth]{fig/base_anchor.png}
\end{subfigure}
\begin{subfigure}{0.47\linewidth}
\includegraphics[width=\linewidth]{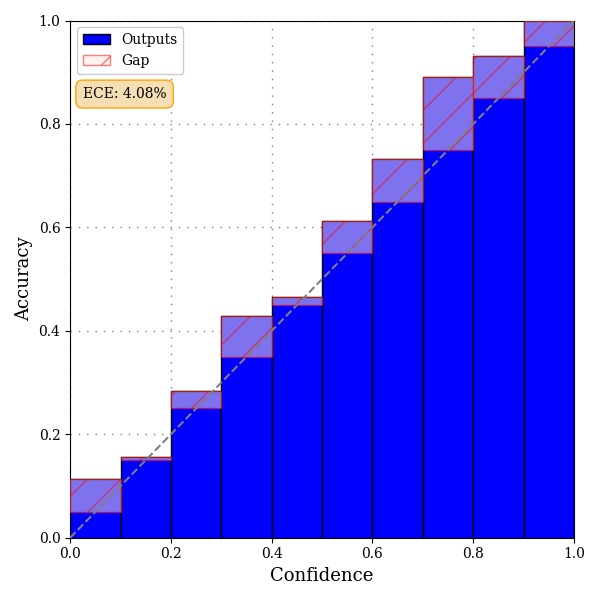}
\end{subfigure}

\vspace{2mm}

\rotatebox[origin=l]{90}{\hspace{3.25em} Novel class}
\begin{subfigure}{0.47\linewidth}
\includegraphics[width=\linewidth]{fig/novel_diversity.png}
\end{subfigure}
\begin{subfigure}{0.47\linewidth}
\includegraphics[width=\linewidth]{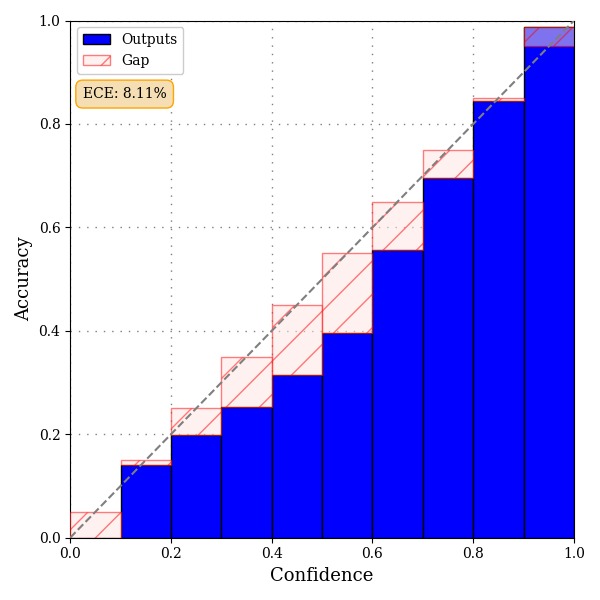}
\end{subfigure}
\caption{$\mathcal{L}_{\text{anchor}}$ analyses on both base and novel classes.}
\label{fig:loss_anchor}
\end{figure}

Fig.~\ref{fig:loss_anchor} illustrates how $\mathcal{L}_{\text{anchor}}$ interacts directly with $\mathcal{L}_{\text{CE}}$ on both base and novel classes in the absence of $\mathcal{L}_{\text{min-max}}$. Applied in isolation, $\mathcal{L}_{\text{anchor}}$
marginally reduces underconfidence on base classes by preserving the pre-trained semantic geometry of CLIP without enforcing class discriminability. Nevertheless, it effectively suppresses overconfidence on novel classes by anchoring tunable embeddings to their corresponding zero-shot embeddings, thereby maintaining the relative class structure of the pre-trained embedding space and preventing semantic drift $(\theta_j\uparrow)$. As shown in Fig.~\ref{fig:loss_anchor}, incorporating $\mathcal{L}_{\text{min-max}}$ alongside $\mathcal{L}_{\text{anchor}}$ further reduces the dual miscalibration problem on both base and novel classes by simultaneously enforcing class discriminability $(\theta_{jk}\uparrow)$ and preserving the pre-trained semantic geometry. Together, these two complementary objectives address the geometric root causes of the dual miscalibration problem, \emph{embedding collapse} and \emph{semantic drift}, yielding well-separated, adjusted decision boundaries that consistently reduce underconfidence on base classes and overconfidence on novel classes.

Tab.~\ref{tab:gradient} provides a detailed dataset-wise ablation of the loss components, extending the average findings reported in the main paper across 10 diverse datasets.

The addition of \(\mathcal{L}_{\text{min-max}}\) to the cross-entropy baseline consistently improves base class calibration. For instance, on OxfordPets, ECE drops from \(2.68\) to \(1.68\), and on Caltech101, it significantly decreases from \(1.54\) to \(0.67\). While this confirms improved discriminability for base classes, novel-class ECE often remains high or increases (e.g., from \(2.21\) to \(5.45\) on SUN397), highlighting overconfidence in novel classes. Integrating \(\mathcal{L}_{\text{anchor}}\) is pivotal for novel class performance. On SUN397, novel ECE drastically improves from \(5.45\) to \(1.02\) (w/o gradient blocking). Similarly, EuroSAT shows a reduction in novel ECE from \(9.30\) to \(4.92\). These results validate that anchoring zero-shot embeddings prevents the distortion typically caused by unconstrained angular maximization. The inclusion of gradient blocking yields the most stable results across both categories. In the DTD base class, ECE improves from \(3.12\) to \(2.14\) when gradient blocked. Crucially, it prevents the collapse of class discriminability, on Aircraft, the full M-PFT objective maintains a competitive novel ECE of \(8.70\), whereas baseline cross-entropy suffers at \(11.61\). Across all individual datasets, the full M-PFT objective (with gradient blocking) achieves the most robust calibration-accuracy trade-off, demonstrating that the synergy between \(\mathcal{L}_{\text{min-max}}\) and \(\mathcal{L}_{\text{anchor}}\) is a consistent requirement regardless of the specific data domain.

\section{Impact of hyperparameters} 
\label{sec:hyperparameter}

For M-PFT, we introduce two regularization coefficients: $\lambda_1$ for min-max regularization ($\mathcal{L}_{\text{min-mx}}$) and 
$\lambda_2$ for zero-shot anchoring regularization ( $\mathcal{L}_{\text{anchor}}$). To determine their optimal values, we perform a sensitivity analysis by sweeping each coefficient independently while holding the other fixed.
As shown in Tab.~\ref{tab:lambda}, we evaluate the impact of these parameters on model accuracy and Expected Calibration Error (ECE), averaged across the Food101, Caltech, and Oxford Flowers datasets. We observe that setting $\lambda_1=0.4$ and $\lambda_2=1.4$ yields a superior balance between base/novel class accuracy and calibration performance (achieving an average Accuracy of \(94.64\) and a low average ECE of \(1.48\)). To ensure a consistent and fair evaluation setting, these heuristically determined values are held fixed across all datasets and prompt-tuning backbones. All reported results are averaged over three random seeds.

\section{Results on different backbones}
\label{sec:backbone}

To assess the generalizability and adaptability of the M-PFT, we evaluate calibration performance when applied to CoOp across two different backbones: ResNet-50 and ViT-B/32. As shown in Tab.~\ref{tab:rn} and Tab.~\ref{tab:b36}, M-PFT demonstrates robust performance across varying backbones, effectively reducing calibration error while maintaining accuracy.

On the ResNet-50 backbone, M-PFT achieves a significant reduction in ECE, reaching an average of 3.78, compared to the 4.04 reported for the baseline. This trend remains consistent for the ViT-B/32 backbone, where M-PFT attains an average ECE of 2.98, notably outperforming the baseline’s 3.15. These results indicate that M-PFT consistently improves the reliability of model predictions.

M-PFT also exhibits strong competitive performance in zero-shot scenarios for novel classes. In ResNet-50, M-PFT achieved the lowest average ECE, down from 9.38 to 4.96, compared to TCPT at 5.46. In ViT-B/32, M-PFT lowers ECE from 10.13 to 4.70, compared to TCPT 5.82. Overall, these findings suggest that M-PFT provides a versatile solution for model calibration that scales effectively across different model backbones without sacrificing discriminative power.

\section{Different prompt initializations}\label{sec:prompt}

We assess robustness using four diverse prompt templates that reflect different context and semantic variations found in prompt engineering literature, \textbf{P1:} ``{\tt a nice image of a [class]}'', \textbf{P2:} ``{\tt an example of [class]}'', \textbf{P3:} ``{\tt a picture of a [class]}'', and \textbf{P4:} ``{\tt a photo of the cool [class]}''. 
Such robustness is valuable for real-world prompt-based VLM deployment, where the most effective prompt template may be unknown in advance or chosen heuristically.

\section{Computational overhead}
\label{sec:overhead}

To assess the overhead of M-PFT, we compared per-batch computation time, GPU memory usage, and ECE against MaPLe~\cite{khattak2023maple} on FGVC Aircraft~\cite{maji2013fine} (16-shot, ViT-B/16) using PyTorch~\cite{paszke2019pytorch}.
M-PFT's average per-batch training time is \(0.057\)\,s versus \(0.055\)\,s for the MaPLe Baseline, an increase of under \(2\)\,ms, and GPU memory usage remains virtually unchanged, while ECE decreases from \(4.77\) to \(3.76\).
These results confirm that M-PFT improves calibration with negligible computational overhead.

\section{Standard deviation across random seeds} 

\label{sec:std}

Tab.~\ref{tab:std} presents the standard deviation of accuracy and Expected Calibration Error (ECE) over three random runs with different seeds for M-PFT across \(10\) datasets for base and novel classes with MMRL~\cite{guo2025mmrl} as the backbone. Compared to the Baseline, M-PFT generally exhibits a lower average standard deviation across both metrics. Specifically, for Base classes, M-PFT achieves an average standard deviation of \(0.26\) for ECE (compared to \(0.29\) for Baseline) and a competitive \(0.35\) for Accuracy. In the Novel class setting, M-PFT shows a marked improvement in stability, reducing the average standard deviation of ECE to \(0.42\) from the Baseline's \(0.47\). These results indicate more stable performance with respect to randomness in prompt initialization and greater consistency in calibration. This consistency highlights M-PFT’s reliability in maintaining stable performance across runs, a critical property for practical deployment in scenarios that demand reproducibility and calibration stability.

\begin{table*}[t!]
\vspace{-10pt}
\centering
\resizebox{\textwidth}{!}{%
\begin{tabular}{@{}l|c|c|cccccccccc|>{\columncolor{pink!20}}r@{}}
\toprule
\textbf{Method} & \textbf{Calibration} & \textbf{Metric} & \rotatebox[origin=c]{45}{\parbox{15mm}{\centering \textbf{DTD}}} & \rotatebox[origin=c]{45}{\parbox{15mm}{\centering \textbf{Flowers102}}} & \rotatebox[origin=c]{45}{\parbox{15mm}{\centering \textbf{Food101}}} & \rotatebox[origin=c]{45}{\parbox{15mm}{\centering \textbf{SUN397}}} & \rotatebox[origin=c]{45}{\parbox{15mm}{\centering \textbf{Aircrafts}}} & \rotatebox[origin=c]{45}{\parbox{15mm}{\centering \textbf{OxfordPets}}} & \rotatebox[origin=c]{45}{\parbox{15mm}{\centering \textbf{Caltech101}}} & \rotatebox[origin=c]{45}{\parbox{15mm}{\centering \textbf{UCF101}}} & \rotatebox[origin=c]{45}{\parbox{15mm}{\centering \textbf{EuroSAT}}} & \rotatebox[origin=c]{45}{\parbox{15mm}{\centering \textbf{Stanford Cars}}} & \rotatebox[origin=c]{45}{\parbox{15mm}{\centering \textit{Average}}} \\
\midrule\midrule

\multirow{2}{*}{\textbf{Baseline}} & \multirow{2}{*}{Zero-Shot}
& Acc. $\uparrow$ & 53.00 & 71.80 & 90.10 & 69.40 & 27.70 & 91.30 & 97.20 & 71.00 & 57.00 & 63.60 & 69.21 \\

& & ECE $\downarrow$ & 4.53 & 3.11 & 1.57 & 1.59 & 3.03 & 2.25 & 6.49 & 3.24 & 8.35 & 3.74 & 3.79 \\

\midrule

\multirow{7}{*}{\textbf{CoOp}} & \multirow{2}{*}{Baseline}
& Acc. $\uparrow$ & 76.27 & 90.00 & 90.20 & 81.14 & 35.23 & 94.77 & 97.98 & 83.32 & 90.24 & 76.22 & 81.54 \\

& & ECE $\downarrow$ & 12.17 & 4.93 & 3.66 & 8.11 & 25.70 & 1.00 & 0.66 & 6.44 & 1.75 & 3.73 & 6.82 \\

\cline{2-14}

& \multirow{2}{*}{TCPT} & Acc. $\uparrow$ & 80.44 & 97.21 & 90.38 & 81.68 & 38.62 & 94.95 & 98.06 & 84.68 & 88.56 & 77.32 & 83.19 \\

& & ECE $\downarrow$ & 2.42 & 4.80 & 0.30 & 1.22 & 4.96 & 1.94 & 1.01 & 1.11 & 4.90 & 7.10 & 2.98 \\
\bottomrule

\end{tabular}%
}
\vspace{-5pt}
\caption{Comparison of calibration performance on \textbf{base classes} across 10 fine-grained classification datasets using \textbf{CLIP} backbone for top@1 Accuracy (Acc.) and Expected Calibration Error (ECE) with multiple prompt-tuning methods across various calibration baselines, where higher Acc and lower ECE indicate better performance and calibration, respectively}
\label{tab:base}
\vspace{-10pt}
\end{table*}

\begin{table*}[t!]
\vspace{-10pt}
\centering
\resizebox{\textwidth}{!}{%
\begin{tabular}{@{}l|c|c|cccccccccc|>{\columncolor{pink!20}}r@{}}
\toprule
\textbf{Method} & \textbf{Calibration} & \textbf{Metric} & \rotatebox[origin=c]{45}{\parbox{15mm}{\centering \textbf{DTD}}} & \rotatebox[origin=c]{45}{\parbox{15mm}{\centering \textbf{Flowers102}}} & \rotatebox[origin=c]{45}{\parbox{15mm}{\centering \textbf{Food101}}} & \rotatebox[origin=c]{45}{\parbox{15mm}{\centering \textbf{SUN397}}} & \rotatebox[origin=c]{45}{\parbox{15mm}{\centering \textbf{Aircrafts}}} & \rotatebox[origin=c]{45}{\parbox{15mm}{\centering \textbf{OxfordPets}}} & \rotatebox[origin=c]{45}{\parbox{15mm}{\centering \textbf{Caltech101}}} & \rotatebox[origin=c]{45}{\parbox{15mm}{\centering \textbf{UCF101}}} & \rotatebox[origin=c]{45}{\parbox{15mm}{\centering \textbf{EuroSAT}}} & \rotatebox[origin=c]{45}{\parbox{15mm}{\centering \textbf{Stanford Cars}}} & \rotatebox[origin=c]{45}{\parbox{15mm}{\centering \textit{Average}}} \\
\midrule\midrule

\multirow{2}{*}{\textbf{Baseline}} & \multirow{2}{*}{Zero-Shot}
& Acc. $\uparrow$ & 60.60 & 77.50 & 91.10 & 75.50 & 35.90 & 97.10 & 94.10 & 78.60 & 63.80 & 75.00 & 74.92 \\

& & ECE $\downarrow$ & 6.86 & 4.91 & 1.83 & 3.48 & 6.55 & 3.42 & 1.55 & 5.52 & 9.12 & 3.31 & 4.65 \\

\midrule

\multirow{8}{*}{\textbf{CoOp}}& \multirow{2}{*}{Baseline}
& Acc. $\uparrow$ & 52.98 & 69.90 & 90.57 & 70.77 & 24.79 & 96.49 & 94.18 & 62.83 & 64.68 & 65.29 & 69.25\\

& & ECE $\downarrow$ & 20.82 & 12.13 & 3.03 & 13.70 & 30.44 & 1.67 & 2.16 & 18.74 & 11.88 & 11.73 & 12.63 \\

\cline{2-14}

& DAC & ECE $\downarrow$ & 10.48 & 10.19 & 1.78 & 4.05 & 17.38 & 1.82 & 3.17 & 8.67 & 8.62 & 5.16 & 7.13 \\

\cline{2-14}

& \multirow{2}{*}{TCPT} & Acc. $\uparrow$ & 48.07 & 71.63 & 90.84 & 70.03 & 30.83 & 97.36 & 93.56 & 66.49 & 56.70 & 69.49 & 69.50 \\

& & ECE $\downarrow$ & 9.31 & 3.51 & 0.87 & 3.08 & 10.64 & 3.03 & 2.21 & 4.75 & 11.15 & 2.10 & 5.06 \\
\bottomrule
\end{tabular}%
}
\end{table*}

\begin{table*}[t!]
\centering
\resizebox{\textwidth}{!}{%
\begin{tabular}{@{}l|c|c|cccccccccc|>{\columncolor{pink!20}}r@{}}
\toprule
\textbf{Method} & \textbf{Calibration} & \textbf{Metric} & \rotatebox[origin=c]{45}{\parbox{15mm}{\centering \textbf{DTD}}} & \rotatebox[origin=c]{45}{\parbox{15mm}{\centering \textbf{Flowers102}}} & \rotatebox[origin=c]{45}{\parbox{15mm}{\centering \textbf{Food101}}} & \rotatebox[origin=c]{45}{\parbox{15mm}{\centering \textbf{SUN397}}} & \rotatebox[origin=c]{45}{\parbox{15mm}{\centering \textbf{Aircrafts}}} & \rotatebox[origin=c]{45}{\parbox{15mm}{\centering \textbf{OxfordPets}}} & \rotatebox[origin=c]{45}{\parbox{15mm}{\centering \textbf{Caltech101}}} & \rotatebox[origin=c]{45}{\parbox{15mm}{\centering \textbf{UCF101}}} & \rotatebox[origin=c]{45}{\parbox{15mm}{\centering \textbf{EuroSAT}}} & \rotatebox[origin=c]{45}{\parbox{15mm}{\centering \textbf{Stanford Cars}}} & \rotatebox[origin=c]{45}{\parbox{15mm}{\centering \textit{Average}}} \\
\midrule\midrule

\multirow{8}{*}{\textbf{HiCroPL}} 
& \multirow{2}{*}{Baseline}
& Acc. $\uparrow$ & 65.14 & 76.15 & 91.88 & 78.16 & 39.47 & 97.49 & 95.52 & 82.03 & 74.45 & 74.92 & 77.52 \\

& & ECE $\downarrow$ & 5.33 & 5.34 & 0.87 & 1.14 & 7.76 & 2.72 & 1.01 & 2.03 & 8.81 & 1.90 & 3.69 \\

\cline{2-14}

& DAC & ECE $\downarrow$  & 4.16 & 6.04 & 1.33 & 1.41 & 7.68 & 2.18 & 1.10 & 3.15 & 8.24 & 2.26 & 3.75 \\

\cline{2-14}

& \multirow{2}{*}{TCPT}
& Acc. $\uparrow$ & 66.51 & 76.10 & 91.79 & 77.37 & 38.89 & 97.69 & 95.20 & 82.39 & 74.74 & 75.21 & 77.59 \\

& & ECE $\downarrow$ & 4.95 & 7.53 & 0.50 & 2.07 & 7.99 & 1.49 & 1.12 & 1.73 & 10.83 & 2.11 & 4.03 \\

\cline{2-14}

& \cellcolor{pink!20}
& \cellcolor{pink!20} \textbf{Acc. $\uparrow$} & \cellcolor{pink!20} 67.79 & \cellcolor{pink!20} 76.50 & \cellcolor{pink!20} 91.80 & \cellcolor{pink!20} 78.15 & \cellcolor{pink!20} 38.33 & \cellcolor{pink!20} 97.54 & \cellcolor{pink!20} 95.74 & \cellcolor{pink!20} 82.28 & \cellcolor{pink!20} 74.60 & \cellcolor{pink!20} 75.35 & 77.81 \\
\bottomrule

\end{tabular}%
}
\vspace{-5pt}
\caption{Comparison of calibration performance on \textbf{novel classes} across 10 fine-grained classification datasets  using \textbf{CLIP} backbone for top@1 Accuracy (Acc.) and Expected Calibration Error (ECE) with multiple prompt-tuning methods across various calibration baselines, where higher Acc and lower ECE indicate better performance and calibration, respectively}
\label{tab:novel}
\vspace{-10pt}
\end{table*}

\begin{table*}[t!]
\centering
\resizebox{\textwidth}{!}{%
\begin{tabular}{@{}l|c|c|cccccccccc|>{\columncolor{pink!20}}r@{}}
\toprule
\textbf{Method} & \textbf{Calibration} & \textbf{Metric} & 
\rotatebox[origin=c]{45}{\parbox{15mm}{\centering \textbf{DTD}}} & 
\rotatebox[origin=c]{45}{\parbox{15mm}{\centering \textbf{Flowers102}}} & 
\rotatebox[origin=c]{45}{\parbox{15mm}{\centering \textbf{Food101}}} & 
\rotatebox[origin=c]{45}{\parbox{15mm}{\centering \textbf{SUN397}}} & 
\rotatebox[origin=c]{45}{\parbox{15mm}{\centering \textbf{Aircrafts}}} & 
\rotatebox[origin=c]{45}{\parbox{15mm}{\centering \textbf{OxfordPets}}} & 
\rotatebox[origin=c]{45}{\parbox{15mm}{\centering \textbf{Caltech101}}} & 
\rotatebox[origin=c]{45}{\parbox{15mm}{\centering \textbf{UCF101}}} & 
\rotatebox[origin=c]{45}{\parbox{15mm}{\centering \textbf{EuroSAT}}} & 
\rotatebox[origin=c]{45}{\parbox{15mm}{\centering \textbf{Stanford Cars}}} & 
\rotatebox[origin=c]{45}{\parbox{15mm}{\centering \textit{Average}}} \\
\midrule\midrule

\multirow{2}{*}{\textbf{Baseline}} & \multirow{2}{*}{Zero-Shot}
& Acc. $\uparrow$ & 76.16 & 91.17 & 92.28 & 75.24 & 29.05 & 96.49 & 98.52 & 75.75 & 57.05 & 85.01 & 77.67 \\
& & ECE $\downarrow$ & 3.80 & 5.07 & 0.61 & 3.81 & 5.81 & 2.07 & 1.00 & 4.76 & 18.42 & 1.07 & 4.64 \\
\midrule

\midrule

\multirow{7}{*}{\textbf{CoOp}} 
& \multirow{2}{*}{Baseline}
& Acc. $\uparrow$ & 88.39 & 98.86 & 92.42 & 82.13 & 34.19 & 95.78 & 98.97 & 83.61 & 90.20 & 88.22 & 85.28 \\
& & ECE $\downarrow$ & 2.36 & 2.02 & 1.45 & 0.92 & 4.38 & 1.79 & 1.25 & 2.26 & 4.97 & 3.99 & 2.54 \\

\cline{2-14}

& \multirow{2}{*}{TCPT}
& Acc. $\uparrow$ & 88.62 & 99.05 & 92.48 & 82.12 & 35.20 & 95.94 & 98.95 & 83.63 & 90.14 & 88.67 & 85.48 \\
\bottomrule
\end{tabular}%
}
\end{table*}

\begin{table*}[t!]
\centering
\resizebox{\textwidth}{!}{%
\begin{tabular}{@{}l|c|c|cccccccccc|>{\columncolor{pink!20}}r@{}}
\toprule
\textbf{Method} & \textbf{Calibration} & \textbf{Metric} & 
\rotatebox[origin=c]{45}{\parbox{15mm}{\centering \textbf{DTD}}} & 
\rotatebox[origin=c]{45}{\parbox{15mm}{\centering \textbf{Flowers102}}} & 
\rotatebox[origin=c]{45}{\parbox{15mm}{\centering \textbf{Food101}}} & 
\rotatebox[origin=c]{45}{\parbox{15mm}{\centering \textbf{SUN397}}} & 
\rotatebox[origin=c]{45}{\parbox{15mm}{\centering \textbf{Aircrafts}}} & 
\rotatebox[origin=c]{45}{\parbox{15mm}{\centering \textbf{OxfordPets}}} & 
\rotatebox[origin=c]{45}{\parbox{15mm}{\centering \textbf{Caltech101}}} & 
\rotatebox[origin=c]{45}{\parbox{15mm}{\centering \textbf{UCF101}}} & 
\rotatebox[origin=c]{45}{\parbox{15mm}{\centering \textbf{EuroSAT}}} & 
\rotatebox[origin=c]{45}{\parbox{15mm}{\centering \textbf{Stanford Cars}}} & 
\rotatebox[origin=c]{45}{\parbox{15mm}{\centering \textit{Average}}} \\
\midrule\midrule

\multirow{7}{*}{\textbf{ProGrad}} 
& \multirow{2}{*}{Baseline}
& Acc. $\uparrow$ & 85.73 & 96.96 & 92.84 & 81.63 & 39.10 & 96.69 & 98.79 & 82.69 & 78.16 & 88.11 & 84.07 \\
& & ECE $\downarrow$ & 9.38 & 6.29 & 1.70 & 3.04 & 5.87 & 2.44 & 1.21 & 3.02 & 4.23 & 3.99 & 4.12 \\

\cline{2-14}

& \multirow{2}{*}{TCPT}
& Acc. $\uparrow$ & 86.92 & 97.57 & 92.89 & 82.10 & 38.46 & 96.46 & 98.95 & 83.18 & 81.02 & 88.09 & 84.56 \\
& & ECE $\downarrow$ & 9.14 & 5.05 & 0.74 & 2.20 & 6.45 & 1.61 & 0.74 & 2.43 & 4.49 & 3.31 & 3.62 \\

\cline{2-14}

& \cellcolor{pink!20} & \cellcolor{pink!20} \textbf{Acc. $\uparrow$} & \cellcolor{pink!20} 85.45 & \cellcolor{pink!20} 96.90 & \cellcolor{pink!20} 92.94 & \cellcolor{pink!20} 81.20 & \cellcolor{pink!20} 36.51 & \cellcolor{pink!20} 96.71 & \cellcolor{pink!20} 98.77 & \cellcolor{pink!20} 81.92 & \cellcolor{pink!20} 78.37 & \cellcolor{pink!20} 87.54 & 83.63 \\
\bottomrule
\end{tabular}%
}
\caption{Comparison of calibration performance on \textbf{base classes} across 10 fine-grained classification datasets using \textbf{SigLIP} backbone for top@1 Accuracy (Acc.) and Expected Calibration Error (ECE) with multiple prompt-tuning methods, where higher Acc and lower ECE indicate better performance and calibration, respectively.}
\label{tab:base_siglip}
\end{table*}

\begin{table*}[t!]
\vspace{-10pt}
\centering
\resizebox{\textwidth}{!}{%
\begin{tabular}{@{}l|c|c|cccccccccc|>{\columncolor{pink!20}}r@{}}
\toprule
\textbf{Method} & \textbf{Calibration} & \textbf{Metric} & 
\rotatebox[origin=c]{45}{\parbox{15mm}{\centering \textbf{DTD}}} & 
\rotatebox[origin=c]{45}{\parbox{15mm}{\centering \textbf{Flowers102}}} & 
\rotatebox[origin=c]{45}{\parbox{15mm}{\centering \textbf{Food101}}} & 
\rotatebox[origin=c]{45}{\parbox{15mm}{\centering \textbf{SUN397}}} & 
\rotatebox[origin=c]{45}{\parbox{15mm}{\centering \textbf{Aircrafts}}} & 
\rotatebox[origin=c]{45}{\parbox{15mm}{\centering \textbf{OxfordPets}}} & 
\rotatebox[origin=c]{45}{\parbox{15mm}{\centering \textbf{Caltech101}}} & 
\rotatebox[origin=c]{45}{\parbox{15mm}{\centering \textbf{UCF101}}} & 
\rotatebox[origin=c]{45}{\parbox{15mm}{\centering \textbf{EuroSAT}}} & 
\rotatebox[origin=c]{45}{\parbox{15mm}{\centering \textbf{Stanford Cars}}} & 
\rotatebox[origin=c]{45}{\parbox{15mm}{\centering \textit{Average}}} \\
\midrule\midrule

\multirow{2}{*}{\textbf{Baseline}} & \multirow{2}{*}{Zero-Shot}
& Acc. $\uparrow$ & 74.88 & 85.67 & 93.59 & 78.57 & 40.85 & 98.15 & 97.82 & 83.78 & 67.21 & 96.21 & 81.67 \\
& & ECE $\downarrow$ & 3.10 & 3.98 & 0.96 & 2.78 & 12.00 & 0.97 & 0.83 & 2.76 & 4.11 & 0.69 & 3.22 \\

\midrule

\multirow{7}{*}{\textbf{CoOp}} 
& \multirow{2}{*}{Baseline}
& Acc. $\uparrow$ & 67.47 & 81.04 & 91.48 & 77.49 & 30.90 & 97.80 & 97.05 & 75.12 & 48.78 & 94.46 & 76.16 \\

& & ECE $\downarrow$ & 13.28 & 8.48 & 0.98 & 3.82 & 17.82 & 1.24 & 1.14 & 3.77 & 16.90 & 1.93 & 6.94 \\

\cline{2-14}

& \multirow{2}{*}{TCPT}
& Acc. $\uparrow$ & 67.75 & 81.09 & 91.85 & 77.46 & 33.44 & 97.50 & 97.71 & 74.94 & 42.16 & 94.26 & 75.82 \\
\bottomrule
\end{tabular}%
}
\end{table*}

\begin{table*}[t!]
\centering
\resizebox{\textwidth}{!}{%
\begin{tabular}{@{}l|c|c|cccccccccc|>{\columncolor{pink!20}}r@{}}
\toprule
\textbf{Method} & \textbf{Calibration} & \textbf{Metric} & 
\rotatebox[origin=c]{45}{\parbox{15mm}{\centering \textbf{DTD}}} & 
\rotatebox[origin=c]{45}{\parbox{15mm}{\centering \textbf{Flowers102}}} & 
\rotatebox[origin=c]{45}{\parbox{15mm}{\centering \textbf{Food101}}} & 
\rotatebox[origin=c]{45}{\parbox{15mm}{\centering \textbf{SUN397}}} & 
\rotatebox[origin=c]{45}{\parbox{15mm}{\centering \textbf{Aircrafts}}} & 
\rotatebox[origin=c]{45}{\parbox{15mm}{\centering \textbf{OxfordPets}}} & 
\rotatebox[origin=c]{45}{\parbox{15mm}{\centering \textbf{Caltech101}}} & 
\rotatebox[origin=c]{45}{\parbox{15mm}{\centering \textbf{UCF101}}} & 
\rotatebox[origin=c]{45}{\parbox{15mm}{\centering \textbf{EuroSAT}}} & 
\rotatebox[origin=c]{45}{\parbox{15mm}{\centering \textbf{Stanford Cars}}} & 
\rotatebox[origin=c]{45}{\parbox{15mm}{\centering \textit{Average}}} \\
\midrule\midrule

\multirow{7}{*}{\textbf{ProDA}} 
& \multirow{2}{*}{Baseline}
& Acc. $\uparrow$ & 70.25 & 82.84 & 91.99 & 80.52 & 39.27 & 97.85 & 97.24 & 77.82 & 49.94 & 95.86 & 78.36 \\
& & ECE $\downarrow$ & 7.91 & 6.34 & 1.63 & 1.68 & 8.48 & 1.45 & 0.79 & 2.17 & 13.87 & 1.95 & 4.63 \\

\cline{2-14}

& \multirow{2}{*}{TCPT}
& Acc. $\uparrow$ & 74.76 & 83.78 & 93.33 & 80.67 & 40.89 & 98.30 & 98.07 & 81.68 & 46.88 & 96.31 & 79.47 \\
& & ECE $\downarrow$ & 3.32 & 5.76 & 0.85 & 0.76 & 5.18 & 1.15 & 0.78 & 2.46 & 21.39 & 1.30 & 4.30 \\

\cline{2-14}

& \cellcolor{pink!20} & \cellcolor{pink!20} \textbf{Acc. $\uparrow$} & \cellcolor{pink!20} 70.49 & \cellcolor{pink!20} 83.60 & \cellcolor{pink!20} 92.98 & \cellcolor{pink!20} 80.91 & \cellcolor{pink!20} 39.33 & \cellcolor{pink!20} 98.21 & \cellcolor{pink!20} 98.00 & \cellcolor{pink!20} 77.47 & \cellcolor{pink!20} 49.79 & \cellcolor{pink!20} 96.23 & 78.70 \\
\bottomrule
\end{tabular}%
}
\vspace{-5pt}
\caption{Comparison of calibration performance on \textbf{novel classes} across 10 fine-grained classification datasets using \textbf{SigLIP} backbone for top@1 Accuracy (Acc.) and Expected Calibration Error (ECE) with multiple prompt-tuning methods, where higher Acc and lower ECE indicate better performance and calibration, respectively.}
\label{tab:novel_siglip}
\vspace{-10pt}
\end{table*}

\begin{table*}[t!]
\centering
\resizebox{0.6\textwidth}{!}{%
\begin{tabular}{@{}l|c|c|ccccc|>{\columncolor{pink!20}}r@{}}
\toprule
\textbf{Class} & \textbf{Calibration} & \textbf{Metric} & \rotatebox[origin=c]{45}{\parbox{15mm}{\centering \textbf{ImageNet-1k}}} & \rotatebox[origin=c]{45}{\parbox{15mm}{\centering \textbf{ImageNet-V2}}} & \rotatebox[origin=c]{45}{\parbox{15mm}{\centering \textbf{ImageNet-S}}} & \rotatebox[origin=c]{45}{\parbox{15mm}{\centering \textbf{ImageNet-A}}} & \rotatebox[origin=c]{45}{\parbox{15mm}{\centering \textbf{ImageNet-R}}} & \rotatebox[origin=c]{45}{\parbox{15mm}{\centering \textit{Average}}} \\
\midrule\midrule
\multirow{9}{*}{\textbf{Base}} & \multirow{2}{*}{Zero-Shot} & Acc. $\uparrow$ & 72.40 & 60.87 & 46.09 & 47.81 & 75.41 & 60.52 \\
& & ECE $\downarrow$ & 1.51 & 2.43 & 4.88 & 8.34 & 2.42 & 3.92 \\
\cline{2-9}
& \multirow{2}{*}{Baseline} & Acc. $\uparrow$ & 76.71 & 67.35 & 53.35 & 68.31 & 85.22 & 70.19 \\
& & ECE $\downarrow$ & 2.27 & 3.14 & 3.94 & 2.52 & 3.14 & 3.00 \\
\cline{2-9}
& \multirow{2}{*}{TCPT} & Acc. $\uparrow$ & 76.72 & 67.19 & 53.15 & 67.86 & 85.28 & 70.04 \\
& & ECE $\downarrow$ & 2.39 & 3.09 & 3.91 & 2.21 & 2.05 & 2.73 \\
\cline{2-9}
& \cellcolor{pink!20} & \cellcolor{pink!20} \textbf{Acc. $\uparrow$} & \cellcolor{pink!20} 76.83 & \cellcolor{pink!20} 66.60 & \cellcolor{pink!20} 58.77 & \cellcolor{pink!20} 67.72 & \cellcolor{pink!20} 86.06 & 71.20 \\
\bottomrule
\end{tabular}%
}
\vspace{-5pt}
\caption{Comparison of calibration performance on both base and novel classes across 4 natural distribution shift datasets for top@1 Accuracy (Acc.) and Expected Calibration Error (ECE) with MaPLe and various calibration baselines, where higher Acc and lower ECE indicate better performance and calibration, respectively.}
\label{tab:ood}
\vspace{-10pt}
\end{table*}

\begin{table*}[t!]
\vspace{-10pt}
\centering
\resizebox{\textwidth}{!}{%
\begin{tabular}{@{}l|c|c|cccccccccc|>{\columncolor{pink!20}}r@{}}
\toprule
\textbf{Method} & \textbf{Calibration} & \textbf{Metric} & \rotatebox[origin=c]{45}{\parbox{15mm}{\centering \textbf{DTD}}} & \rotatebox[origin=c]{45}{\parbox{15mm}{\centering \textbf{Flowers102}}} & \rotatebox[origin=c]{45}{\parbox{15mm}{\centering \textbf{Food101}}} & \rotatebox[origin=c]{45}{\parbox{15mm}{\centering \textbf{SUN397}}} & \rotatebox[origin=c]{45}{\parbox{15mm}{\centering \textbf{Aircrafts}}} & \rotatebox[origin=c]{45}{\parbox{15mm}{\centering \textbf{OxfordPets}}} & \rotatebox[origin=c]{45}{\parbox{15mm}{\centering \textbf{Caltech101}}} & \rotatebox[origin=c]{45}{\parbox{15mm}{\centering \textbf{UCF101}}} & \rotatebox[origin=c]{45}{\parbox{15mm}{\centering \textbf{EuroSAT}}} & \rotatebox[origin=c]{45}{\parbox{15mm}{\centering \textbf{Stanford Cars}}} & \rotatebox[origin=c]{45}{\parbox{15mm}{\centering \textit{Average}}} \\
\midrule\midrule

\multirow{7}{*}{\textbf{MaPLe}} &
\multirow{2}{*}{Baseline} & MCE $\downarrow$ & 0.70 & 0.71 & 0.46 & 0.39 & 0.89 & 0.51 & 0.62 & 0.71 & 1.96 & 0.86 & 0.78 \\
& & ACE $\downarrow$ & 1.96 & 1.31 & 1.37 & 1.35 & 3.04 & 1.44 & 1.15 & 2.13 & 7.68 & 3.31 & 2.47 \\
\cline{2-14}
& \multirow{2}{*}{TCPT} & MCE $\downarrow$ & 0.51 & 0.29 & 0.12 & 0.28 & 0.81 & 0.18 & 0.21 & 0.55 & 1.84 & 0.55 & 0.53 \\
& & ACE $\downarrow$ & 1.36 & 0.71 & 0.34 & 0.80 & 2.68 & 0.46 & 0.34 & 1.60 & 5.98 & 1.48 & 1.57 \\
\cline{2-14}
& \cellcolor{pink!20} & \cellcolor{pink!20} \textbf{MCE $\downarrow$} & \cellcolor{pink!20} 0.64 & \cellcolor{pink!20} 1.16 & \cellcolor{pink!20} 0.27 & \cellcolor{pink!20} 0.25 & \cellcolor{pink!20} 0.69 & \cellcolor{pink!20} 0.57 & \cellcolor{pink!20} 0.34 & \cellcolor{pink!20} 0.60 & \cellcolor{pink!20} 1.20 & \cellcolor{pink!20} 0.60 & 0.63 \\
\bottomrule
\end{tabular}%
}
\vspace{-5pt}
\caption{Comparison of calibration performance on base classes across 10 fine-grained classification datasets using \textbf{SigLIP} backbone for Maximum Calibration Error (MCE) and Adaptive Calibration Error (ACE) with multiple prompt-tuning methods across various calibration baselines, where lower MCE and ACE indicate better performance and calibration.}
\label{tab:base_mce_ace}
\vspace{-10pt}
\end{table*}

\begin{table*}[t!]
\centering
\resizebox{\textwidth}{!}{%
\begin{tabular}{@{}l|c|c|cccccccccc|>{\columncolor{pink!20}}r@{}}
\toprule
\textbf{Method} & \textbf{Calibration} & \textbf{Metric} & \rotatebox[origin=c]{45}{\parbox{15mm}{\centering \textbf{DTD}}} & \rotatebox[origin=c]{45}{\parbox{15mm}{\centering \textbf{Flowers102}}} & \rotatebox[origin=c]{45}{\parbox{15mm}{\centering \textbf{Food101}}} & \rotatebox[origin=c]{45}{\parbox{15mm}{\centering \textbf{SUN397}}} & \rotatebox[origin=c]{45}{\parbox{15mm}{\centering \textbf{Aircrafts}}} & \rotatebox[origin=c]{45}{\parbox{15mm}{\centering \textbf{OxfordPets}}} & \rotatebox[origin=c]{45}{\parbox{15mm}{\centering \textbf{Caltech101}}} & \rotatebox[origin=c]{45}{\parbox{15mm}{\centering \textbf{UCF101}}} & \rotatebox[origin=c]{45}{\parbox{15mm}{\centering \textbf{EuroSAT}}} & \rotatebox[origin=c]{45}{\parbox{15mm}{\centering \textbf{Stanford Cars}}} & \rotatebox[origin=c]{45}{\parbox{15mm}{\centering \textit{Average}}} \\ \midrule\midrule
\multirow{7}{*}{\textbf{MaPLe}} & \multirow{2}{*}{Baseline} & MCE $\downarrow$ & 2.50 & 1.50 & 0.48 & 0.21 & 2.3 & 0.50 & 0.38 & 0.75 & 6.72 & 0.48 & 1.58 \\
& & ACE $\downarrow$ & 7.00 & 6.23 & 1.14 & 0.46 & 10.03 & 0.61 & 1.03 & 2.68 & 14.89 & 1.53 & 4.56 \\ 
\cline{2-14}
& \multirow{2}{*}{TCPT} & MCE $\downarrow$ & 1.57 & 1.46 & 0.16 & 0.38 & 1.68 & 0.31 & 0.28 & 0.73 & 1.76 & 0.25 & 0.86 \\
& & ACE $\downarrow$ & 6.23 & 6.09 & 0.37 & 1.54 & 7.19 & 0.38 & 0.53 & 2.04 & 7.11 & 1.12 & 3.26 \\ 
\cline{2-14}
& \cellcolor{pink!20} & \cellcolor{pink!20} \textbf{MCE $\downarrow$} & \cellcolor{pink!20} 1.65 & \cellcolor{pink!20} 0.85 & \cellcolor{pink!20} 0.41 & \cellcolor{pink!20} 0.30 & \cellcolor{pink!20} 2.28 & \cellcolor{pink!20} 0.50 & \cellcolor{pink!20} 0.22 & \cellcolor{pink!20} 0.83 & \cellcolor{pink!20} 5.28 & \cellcolor{pink!20} 0.29 & 1.26 \\
\bottomrule
\end{tabular}%
}
\vspace{-5pt}
\caption{Comparison of calibration performance on novel classes across 10 fine-grained classification datasets using the \textbf{SigLIP} backbone for Maximum Calibration Error (MCE) and Adaptive Calibration Error (ACE), with multiple prompt-tuning methods and across various calibration baselines, where lower MCE and ACE indicate better calibration performance.}
\label{tab:novel_mce_ace}
\vspace{-10pt}
\end{table*}

\begin{table*}[t!]
\vspace{-10pt}
\centering
\resizebox{\textwidth}{!}{%
\begin{tabular}{@{}l|c|c|cccccccccc|>{\columncolor{pink!20}}r@{}}
\toprule
\textbf{Class} & \textbf{Loss
Component} & \textbf{Metric} & \rotatebox[origin=c]{45}{\parbox{15mm}{\centering \textbf{DTD}}} & \rotatebox[origin=c]{45}{\parbox{15mm}{\centering \textbf{Flowers102}}} & \rotatebox[origin=c]{45}{\parbox{15mm}{\centering \textbf{Food101}}} & \rotatebox[origin=c]{45}{\parbox{15mm}{\centering \textbf{SUN397}}} & \rotatebox[origin=c]{45}{\parbox{15mm}{\centering \textbf{Aircrafts}}} & \rotatebox[origin=c]{45}{\parbox{15mm}{\centering \textbf{OxfordPets}}} & \rotatebox[origin=c]{45}{\parbox{15mm}{\centering \textbf{Caltech101}}} & \rotatebox[origin=c]{45}{\parbox{15mm}{\centering \textbf{UCF101}}} & \rotatebox[origin=c]{45}{\parbox{15mm}{\centering \textbf{EuroSAT}}} & \rotatebox[origin=c]{45}{\parbox{15mm}{\centering \textbf{Stanford Cars}}} & \rotatebox[origin=c]{45}{\parbox{15mm}{\centering \textit{Average}}} \\
\midrule\midrule

\multirow{10}{*}{Base} & \multirow{2}{*}{\(\mathcal{L}_{\text{CE}}\)} & Acc. $\uparrow$ & 79.63 & 96.00 & 90.80 & 80.55 & 36.33 & 95.53 & 97.97 & 83.20 & 91.13 & 72.93 & 82.41 \\

& & ECE $\downarrow$ & 4.18 & 4.28 & 0.78 & 1.27 & 4.77 & 2.68 & 1.54 & 2.68 & 3.42 & 7.25 & 3.29 \\

\cline{2-14}

& \multirow{2}{*}{\(\mathcal{L}_{\text{CE}} + \mathcal{L}_{\text{min-max}}\)} & Acc. $\uparrow$ & 78.47 & 96.01 & 90.07 & 80.12 & 33.13 & 95.64 & 97.61 & 83.00 & 90.67 & 72.70 & 81.74 \\

& & ECE $\downarrow$ & 3.50 & 3.14 & 1.49 & 2.09 & 3.32 & 1.68 & 0.67 & 1.65 & 3.76 & 3.60 & 2.49 \\

\cline{2-14}

& \(\mathcal{L}_{\text{CE}} + \mathcal{L}_{\text{min-max}} + \mathcal{L}_{\text{anchor}}\) & Acc. $\uparrow$ & 79.55 & 96.07 & 90.42 & 80.62 & 36.81 & 95.30 & 97.55 & 84.25 & 91.55 & 72.17 & 82.43 \\

& (w/o gradient blocking) & ECE $\downarrow$ & 2.15 & 4.50 & 0.80 & 1.54 & 3.73 & 2.38 & 1.77 & 2.91 & 6.83 & 4.54 & 3.12 \\
\end{tabular}%

\multirow{10}{*}{Novel} &
\multirow{2}{*}{\(\mathcal{L}_{\text{CE}}\)} & Acc. $\uparrow$ & 58.47 & 72.80 & 92.10 & 78.20 & 34.53 & 97.85 & 95.10 & 77.85 & 75.90 & 73.57 & 75.64 \\

& & ECE $\downarrow$ & 12.16 & 11.67 & 1.19 & 2.21 & 11.61 & 2.63 & 1.62 & 3.98 & 11.68 & 3.09 & 6.18 \\

\cline{2-14}

& \multirow{2}{*}{\(\mathcal{L}_{\text{CE}} + \mathcal{L}_{\text{min-max}}\)} & Acc. $\uparrow$ & 53.42 & 71.82 & 91.00 & 76.68 & 25.27 & 97.76 & 94.25 & 77.82 & 58.02 & 74.58 & 72.06 \\

& & ECE $\downarrow$ & 17.14 & 11.41 & 1.13 & 5.45 & 12.13 & 1.90 & 2.46 & 4.35 & 9.30 & 2.77 & 6.80 \\

\cline{2-14}

& \(\mathcal{L}_{\text{CE}} + \mathcal{L}_{\text{min-max}} + \mathcal{L}_{\text{anchor}}\) & Acc. $\uparrow$ & 59.26 & 75.56 & 91.67 & 78.12 & 36.83 & 97.63 & 94.91 & 79.36 & 70.80 & 74.65 & 75.88 \\

& (w/o gradient blocking) & ECE $\downarrow$ & 7.67 & 9.09 & 1.34 & 1.02 & 8.28 & 2.70 & 1.03 & 2.70 & 4.92 & 2.35 & 4.11 \\
\bottomrule

\end{tabular}%
}
\vspace{-5pt}
\caption{Loss component ablation on the effect of min-max regularization \(\mathcal{L}_{\text{min-max}}\) and anchoring regularization \(\mathcal{L}_{\text{anchor}}\) with (w/) and without (w/o) gradient blocking
on both base and novel classes.}
\label{tab:gradient}
\vspace{-10pt}
\end{table*}

\begin{table*}[t!]
\centering
\resizebox{\textwidth}{!}{%
\begin{tabular}{@{}l|c|c|cccccccccc|>{\columncolor{pink!20}}r@{}}
\toprule
\textbf{Method} & \textbf{Calibration} & \textbf{Metric} & \rotatebox[origin=c]{45}{\parbox{15mm}{\centering \textbf{DTD}}} & \rotatebox[origin=c]{45}{\parbox{15mm}{\centering \textbf{Flowers102}}} & \rotatebox[origin=c]{45}{\parbox{15mm}{\centering \textbf{Food101}}} & \rotatebox[origin=c]{45}{\parbox{15mm}{\centering \textbf{SUN397}}} & \rotatebox[origin=c]{45}{\parbox{15mm}{\centering \textbf{Aircrafts}}} & \rotatebox[origin=c]{45}{\parbox{15mm}{\centering \textbf{OxfordPets}}} & \rotatebox[origin=c]{45}{\parbox{15mm}{\centering \textbf{Caltech101}}} & \rotatebox[origin=c]{45}{\parbox{15mm}{\centering \textbf{UCF101}}} & \rotatebox[origin=c]{45}{\parbox{15mm}{\centering \textbf{EuroSAT}}} & \rotatebox[origin=c]{45}{\parbox{15mm}{\centering \textbf{Stanford Cars}}} & \rotatebox[origin=c]{45}{\parbox{15mm}{\centering \textit{Average}}} \\ \midrule\midrule

\multirow{7}{*}{\textbf{CoOp-ResNet50}} & \multirow{2}{*}{Baseline} & Acc. & 74.85 & 95.25 & 82.54 & 76.37 & 29.95 & 90.5 & 95.22 & 80.4 & 81.32 & 70.22 & 77.66 \\

& & ECE & 8.65 & 4.22 & 1.15 & 2.81 & 2.62 & 2.32 & 1.27 & 1.82 & 9.34 & 6.15 & 4.04 \\

\cline{2-14}

& \multirow{2}{*}{TCPT} & Acc. & 75.42 & 95.22 & 82.77 & 77.07 & 29.01 & 91.08 & 95.44 & 79.96 & 89.79 & 69.90 & 78.57 \\

& & ECE & 5.87 & 4.74 & 0.84 & 1.23 & 3.32 & 2.21 & 1.37 & 2.21 & 4.12 & 8.65 & 3.46 \\

\cline{2-14}

& \cellcolor{pink!20} & \cellcolor{pink!20} \textbf{Acc.} & \cellcolor{pink!20} 75.12 & \cellcolor{pink!20} 94.02 & \cellcolor{pink!20} 84.18 & \cellcolor{pink!20} 77.41 & \cellcolor{pink!20} 28.45 & \cellcolor{pink!20} 91.94 & \cellcolor{pink!20} 95.16 & \cellcolor{pink!20} 79.25 & \cellcolor{pink!20} 87.29 & \cellcolor{pink!20} 68.55 & 78.14 \\
\bottomrule

\end{tabular}%
}
\vspace{-5pt}
\caption{Comparison of calibration performance on base classes across 10 fine-grained classification tasks using ResNet-50 and ViT-B/32. We report top-1 accuracy (Acc.) and Expected Calibration Error (ECE) for CoOp~\cite{zhou2022learning}, a prompt-tuning technique, evaluated with different backbones.}
\label{tab:rn}
\vspace{-10pt}
\end{table*}

\begin{table*}[t!]
\centering
\resizebox{\textwidth}{!}{%
\begin{tabular}{@{}l|c|c|cccccccccc|>{\columncolor{pink!20}}r@{}}
\toprule
\textbf{Method} & \textbf{Calibration} & \textbf{Metric} & \rotatebox[origin=c]{45}{\parbox{15mm}{\centering \textbf{DTD}}} & \rotatebox[origin=c]{45}{\parbox{15mm}{\centering \textbf{Flowers102}}} & \rotatebox[origin=c]{45}{\parbox{15mm}{\centering \textbf{Food101}}} & \rotatebox[origin=c]{45}{\parbox{15mm}{\centering \textbf{SUN397}}} & \rotatebox[origin=c]{45}{\parbox{15mm}{\centering \textbf{Aircrafts}}} & \rotatebox[origin=c]{45}{\parbox{15mm}{\centering \textbf{OxfordPets}}} & \rotatebox[origin=c]{45}{\parbox{15mm}{\centering \textbf{Caltech101}}} & \rotatebox[origin=c]{45}{\parbox{15mm}{\centering \textbf{UCF101}}} & \rotatebox[origin=c]{45}{\parbox{15mm}{\centering \textbf{EuroSAT}}} & \rotatebox[origin=c]{45}{\parbox{15mm}{\centering \textbf{Stanford Cars}}} & \rotatebox[origin=c]{45}{\parbox{15mm}{\centering \textit{Average}}} \\ \midrule\midrule

\multirow{7}{*}{\textbf{CoOp-ResNet50}} & \multirow{2}{*}{Baseline} & Acc. & 41.67 & 61.87 & 82.55 & 64.46 & 18.72 & 92.11 & 87.27 & 55.07 & 34.15 & 57.84 & 59.57 \\

& & ECE & 5.47 & 10.24 & 0.86 & 9.07 & 18.75 & 2.10 & 3.57 & 13.47 & 22.18 & 8.08 & 9.38 \\

\cline{2-14}

& \multirow{2}{*}{TCPT} & Acc. & 39.05 & 61.47 & 82.50 & 64.94 & 21.66 & 93.08 & 87.52 & 53.94 & 42.75 & 59.88 & 60.68 \\

& & ECE & 8.83 & 4.13 & 1.72 & 4.53 & 5.36 & 2.68 & 2.82 & 10.16 & 10.93 & 3.43 & 5.46 \\

\cline{2-14}

& \cellcolor{pink!20} & \cellcolor{pink!20} \textbf{Acc.} & \cellcolor{pink!20} 46.46 & \cellcolor{pink!20} 71.07 & \cellcolor{pink!20} 84.68 & \cellcolor{pink!20} 70.71 & \cellcolor{pink!20} 21.72 & \cellcolor{pink!20} 94.93 & \cellcolor{pink!20} 91.38 & \cellcolor{pink!20} 65.66 & \cellcolor{pink!20} 45.76 & \cellcolor{pink!20} 64.22 & 65.66 \\
\bottomrule

\end{tabular}%
}
\vspace{-5pt}
\caption{Comparison of calibration performance on novel classes across 10 fine-grained classification tasks using ResNet-50 and ViT-B/32. We report top-1 accuracy (Acc.) and Expected Calibration Error (ECE) for CoOp~\cite{zhou2022learning}, a prompt-tuning technique, evaluated with different backbones.}
\label{tab:b36}
\vspace{-10pt}
\end{table*}

\begin{table*}[t]
\centering
\begin{minipage}[t]{0.50\textwidth}
\centering
\resizebox{\linewidth}{!}{%
\begin{tabular}{l|c|cc|>{\columncolor{pink!20}}c|cc|>{\columncolor{pink!20}}r@{}}
\toprule
\multirow{2}{*}{$\mathbf{\lambda_1}$} & \multirow{2}{*}{$\mathbf{\lambda_2}$} & \multicolumn{3}{c|}{\textbf{Acc. $\uparrow$}} & \multicolumn{3}{c}{\textbf{ECE $\downarrow$}} \\
\cline{3-8}
& & \textbf{Base} & \textbf{Novel} & \textbf{Average} & \textbf{Base} & \textbf{Novel} & \textbf{Average} \\ 
\midrule 
0.0 & 1.4 & 96.82 & 92.65 & 94.73 & 1.72 & 1.98 & 1.85 \\ 
0.2 & 1.4 & 96.92 & 92.46 & 94.69 & 1.63 & 1.85 & 1.74 \\ 
\midrule
0.4 & 0.0 & 96.45 & 91.45 & 93.95 & 0.83 & 3.29 & 2.06 \\ 
0.4 & 0.5 & 96.78 & 92.19 & 94.48 & 1.00 & 1.90 & 1.45 \\ 
0.4 & 1.0 & 96.77 & 92.41 & 94.59 & 1.24 & 2.01 & 1.62 \\ 
\midrule
\rowcolor{pink!20} \textbf{0.4} & \textbf{1.4} & 96.73 & 92.55 & 94.64 & 1.28 & 1.69 & 1.48 \\ 
\midrule
0.4 & 2.0 & 96.83 & 92.46 & 94.64 & 1.41 & 2.21 & 1.81 \\ 
0.4 & 3.0 & 96.79 & 92.65 & 94.72 & 1.62 & 1.92 & 1.77 \\ 
\midrule
0.6 & 1.4 & 96.74 & 92.29 & 94.52 & 1.23 & 1.98 & 1.60 \\ 
0.8 & 1.4 & 96.74 & 92.49 & 94.61 & 1.11 & 1.86 & 1.49 \\ 
1.0 & 1.4 & 96.68 & 92.48 & 94.58 & 1.02 & 1.72 & 1.37 \\ 
\bottomrule
\end{tabular}%
}
\caption{Hyperparameter tuning for $\lambda_1$ (min-max) and $\lambda_2$ (anchor), averaged across Food101, Caltech, and Oxford Flowers, reported on both base and novel classes.}
\label{tab:lambda}
\end{minipage}%
\hfill
\begin{minipage}[t]{0.46\textwidth}
\centering
\begin{tabular}{l|cc}
\toprule
\multirow{2}{*}{\textbf{Method}} & \textbf{Time (s)} & \textbf{Memory} \\
& \textbf{/ Batch} & \textbf{(MiB)} \\
\midrule
Baseline & 0.055 & 2078 \\
\midrule
\rowcolor{pink!20} \textbf{M-PFT (Ours)} & 0.057 & 2078 \\
\bottomrule
\end{tabular}
\caption{Calculation time per batch (in seconds) and occupied memory usage.}
\label{tab:overhead}
\end{minipage}
\end{table*}

\begin{table*}[t!]
\centering
\resizebox{\textwidth}{!}{%
\begin{tabular}{@{}l|c|c|cccccccccc|>{\columncolor{pink!20}}r@{}}
\toprule
\textbf{Method} & \textbf{Calibration} & \textbf{Metric} & \rotatebox[origin=c]{45}{\parbox{15mm}{\centering \textbf{DTD}}} & \rotatebox[origin=c]{45}{\parbox{15mm}{\centering \textbf{Flowers102}}} & \rotatebox[origin=c]{45}{\parbox{15mm}{\centering \textbf{Food101}}} & \rotatebox[origin=c]{45}{\parbox{15mm}{\centering \textbf{SUN397}}} & \rotatebox[origin=c]{45}{\parbox{15mm}{\centering \textbf{Aircrafts}}} & \rotatebox[origin=c]{45}{\parbox{15mm}{\centering \textbf{OxfordPets}}} & \rotatebox[origin=c]{45}{\parbox{15mm}{\centering \textbf{Caltech101}}} & \rotatebox[origin=c]{45}{\parbox{15mm}{\centering \textbf{UCF101}}} & \rotatebox[origin=c]{45}{\parbox{15mm}{\centering \textbf{EuroSAT}}} & \rotatebox[origin=c]{45}{\parbox{15mm}{\centering \textbf{Stanford Cars}}} & \rotatebox[origin=c]{45}{\parbox{15mm}{\centering \textit{Average}}} \\ \midrule\midrule

\multirow{5}{*}{\textbf{Base}} & \multirow{2}{*}{Baseline} & Std. Acc & 0.44 & 0.18 & 0.14 & 0.00 & 0.78 & 0.27 & 0.11 & 0.32 & 0.96 & 0.42 & 0.36 \\

& & Std. ECE & 0.24 & 0.08 & 0.20 & 0.00 & 0.60 & 0.20 & 0.17 & 0.55 & 0.64 & 0.19 & 0.29 \\

\cline{2-14}

& \cellcolor{pink!20} & \cellcolor{pink!20} \textbf{Std. Acc} & \cellcolor{pink!20} 0.45 & \cellcolor{pink!20} 0.18 & \cellcolor{pink!20} 0.14 & \cellcolor{pink!20} 0.00 & \cellcolor{pink!20} 0.79 & \cellcolor{pink!20} 0.28 & \cellcolor{pink!20} 0.12 & \cellcolor{pink!20} 0.25 & \cellcolor{pink!20} 0.84 & \cellcolor{pink!20} 0.43 & 0.35 \\

& \cellcolor{pink!20} \multirow{-2}{*}{\textbf{M-PFT (Ours)}} & \cellcolor{pink!20} \textbf{Std. ECE} & \cellcolor{pink!20} 0.19 & \cellcolor{pink!20} 0.08 & \cellcolor{pink!20} 0.20 & \cellcolor{pink!20} 0.00 & \cellcolor{pink!20} 0.46 & \cellcolor{pink!20} 0.22 & \cellcolor{pink!20} 0.17 & \cellcolor{pink!20} 0.56 & \cellcolor{pink!20} 0.53 & \cellcolor{pink!20} 0.18 & 0.26 \\
\bottomrule

\end{tabular}%
}
\vspace{-5pt}
\caption{Standard deviation (Std.) across \(3\) random seeds for base and novel classes across \(10\) fine-grained classification benchmarks.}
\label{tab:std}
\vspace{-10pt}
\end{table*}




\end{document}